\documentclass[12pt]{article}

\usepackage{amsfonts}
\usepackage{amsmath}
\usepackage{amsthm}
\usepackage{esint}
\usepackage[margin=1in]{geometry}
\usepackage{graphicx}
\usepackage{xcolor}
\usepackage{epstopdf}
\usepackage{amssymb}
\usepackage{cite}
\usepackage{hyperref}
\hypersetup{
  colorlinks   = true, %
  urlcolor     = black, %
  linkcolor    = black, %
  citecolor   = black %
}

\usepackage{caption}
\usepackage{color}
\usepackage{subcaption}

\usepackage{multicol}
\usepackage{multirow}
\usepackage{makecell}
\numberwithin{equation}{section}

\usepackage{algorithm}
\usepackage{algorithmic}

\theoremstyle{definition}

\newcommand{\R}{\mathbb{R}}

\newcommand{\norm}[1]{||#1||}
\newcommand{\fpar}[2]{\frac{\partial #1}{\partial #2}}

\renewcommand{\vec}[1]{\mathbf{#1}}

\usepackage{titlesec}
\usepackage{chngcntr}
\newcommand{\appendixsectionformat}{%
  \titleformat{\section}[block]{\normalfont\Large\bfseries}{Appendix \thesection}{1em}{}}
\newcommand{\appendixfiguretablenumbering}{%
\renewcommand{\thefigure}{\thesection.\arabic{figure}}%
\renewcommand{\thetable}{\thesection.\arabic{table}}%
\counterwithin{figure}{section}%
\counterwithin{table}{section}}

\usepackage{titling}

\usepackage[capitalize,nameinlink]{cleveref}[0.19]

\Crefname{figure}{Figure}{Figures}

\crefformat{equation}{\textup{#2(#1)#3}}
\crefrangeformat{equation}{\textup{#3(#1)#4--#5(#2)#6}}
\crefmultiformat{equation}{\textup{#2(#1)#3}}{ and \textup{#2(#1)#3}}
{, \textup{#2(#1)#3}}{, and \textup{#2(#1)#3}}
\crefrangemultiformat{equation}{\textup{#3(#1)#4--#5(#2)#6}}%
{ and \textup{#3(#1)#4--#5(#2)#6}}{, \textup{#3(#1)#4--#5(#2)#6}}{, and \textup{#3(#1)#4--#5(#2)#6}}

\Crefformat{equation}{#2Equation~\textup{(#1)}#3}
\Crefrangeformat{equation}{Equations~\textup{#3(#1)#4--#5(#2)#6}}
\Crefmultiformat{equation}{Equations~\textup{#2(#1)#3}}{ and \textup{#2(#1)#3}}
{, \textup{#2(#1)#3}}{, and \textup{#2(#1)#3}}
\Crefrangemultiformat{equation}{Equations~\textup{#3(#1)#4--#5(#2)#6}}%
{ and \textup{#3(#1)#4--#5(#2)#6}}{, \textup{#3(#1)#4--#5(#2)#6}}{, and \textup{#3(#1)#4--#5(#2)#6}}

\usepackage{tikz}
\usetikzlibrary{positioning}
\usetikzlibrary{calc}

\title{Dynamics-Encoded Deep Learning for Robust System Identification and Parameter Estimation} 
\author{Caitlin Ho, Andrea Arnold*}
\date{}

\begin{document}
\maketitle
\vspace{-1.0cm}

\small %

\centerline{Department of Mathematical Sciences, Worcester Polytechnic Institute, Worcester, MA, USA}
\vspace{.2cm}

\centerline{* Corresponding author: anarnold@wpi.edu, ORCID \href{https://orcid.org/0000-0003-3003-882X}{0000-0003-3003-882X}}

\normalsize

\bigskip

\begin{abstract}

\noindent Incorporating {a priori} physics knowledge into machine learning leads to more robust and interpretable algorithms. In this work, we combine deep learning techniques and classic numerical methods for differential equations to address two challenging missing physics problems in dynamical systems theory: dynamics discovery and parameter estimation. 
The presented methods encode available information relating to the system dynamics into deep learning architectures, incorporating different assumptions on the known inputs and desired outputs in each case.
Results demonstrate the effectiveness of the proposed approaches in making data-driven model predictions given corrupt system observations on a suite of test problems exhibiting oscillatory and chaotic dynamics. When comparing the performance of various numerical schemes, such as the Runge-Kutta and linear multistep families of methods, we observe promising results in predicting the system dynamics and estimating physical parameters, given appropriate choices of spatial and temporal discretization schemes and numerical method orders. \\


\end{abstract}

\section{Introduction}
\label{sec:intro}
Mathematical modeling often involves examining the key attributes of dynamical systems that can be used to explain and predict the behavior of many real-world phenomena. Two important aspects of this process, system identification and parameter estimation, aim to improve our understanding of the system dynamics by learning and/or characterizing the predictive model equations through use of observed data. Recently, a growing number of proposed methods numerically approximate the system dynamics directly from observed data \cite{Brunton2016SINDy, Rudy2019, Raissi2018HPM}. Dynamics discovery can be considered as a function approximation problem by treating the unknown governing equations as target functions dependent on the system states and time derivatives. This problem has been explored in the context of both ordinary differential equations (ODEs) \cite{Raissi2018LMM,Rudy2019,Tipireddy2022LMM} and partial differential equations (PDEs) \cite{Raissi2019PINN,Raissi2018HPM,Lu2021DeepONet,PINNsurvey}. Furthermore, there has been increased interest in combining scientific knowledge with machine learning (ML) to create more robust and interpretable algorithms in the emerging field of {scientific machine learning} (SciML) \cite{sciml, Karniadakis2021, Cuomo2022, Willard2022SciML,PeNNsurvey2024,Kim2021}.

In this work, we introduce a novel SciML approach that combines deep learning and numerical methods to address system identification and parameter estimation problems in deterministic dynamical systems governed by differential equations. Various approaches in the literature combine scientific knowledge and ML to varying degrees. One type of approach enforces domain knowledge into ML to constrain the set of possible approximate functions \cite{Raissi2019PINN, PINNsurvey, DeepXDE, Rao2021PeNN, daw2021PGNN}. Another perspective utilizes ML to discover new domain knowledge by learning from an undefined system \cite{Brunton2016SINDy, Raissi2018HPM, Churchill2023Missing}. Within this framework, we consider how to combine ML and scientific computing with the goal of improving dynamical system predictions by (i) enforcing domain knowledge in cases where it is known and (ii) augmenting with data and ML in cases where it is unknown. 

\begin{figure}
    \centering
    \resizebox{0.6\linewidth}{!}{
        \begin{tikzpicture}[scale=0.6]
            
            \draw[black] (0,0) -- (10,0) node[right,align=center] {Physics \\ Models};
            \node (y) at (0,5) {};
            \node [rotate=90,left=of y] at (0.5, 5.75) {Data};
            \draw[black] (0,0) -- (0,10) node[above,align=center] {Supervised \\ ML};
            \node (x) at (5, 0) {};
            \node [below=of x] at (5, 0.8) {Domain Knowledge};
            \draw[gray,opacity=0.3] (10,0) -- (0,10);
            \node (penn) at (8.7,1.3) {PeNN \cite{Rao2021PeNN}};
            \node (pgnn) at (2.2,5.9) {PgNN \cite{daw2021PGNN}};
            \node[align=center] (no) at (5.2, 8.2) {Neural operator \cite{Lu2021DeepONet}};
            \draw[dashed, red, rounded corners=3pt] (5.5,2.5) rectangle (10,5) node[pos=.5,align=center] (proposed) {Proposed \\ Methodology};
            \node [left=of proposed, xshift=1cm] (pinn) at (5.5,4) {PINN \cite{Raissi2019PINN}};
            \node [above=of proposed, yshift=1cm,align=center] (node) at (7.7,1.5) {Neural ODE \cite{NODE}};

            \coordinate (topLeft)     at (5.5,5);
            \coordinate (bottomRight) at (10,2.5);
            
            \begin{scope}[shift={(10,8)},scale=2.0]
              \coordinate (zoomTopLeft)     at (0,2.5);
              \coordinate (zoomBottomRight) at (4.5,0);
              
              \draw[red, dashed, rounded corners=3pt] (0,0) rectangle (4.5,2.5)
                node[pos=.5, align=center] {};
            
              \node[align=center] (missing) at (0.9,0.7) {DEDL for \\ dynamics \\ discovery};
              \node[align=center] (inv) at (3.5,1.75) {DEDL for \\ parameter \\ estimation};
            \end{scope}
            
            \draw[thick, red, opacity=0.3]
              (topLeft) -- ($(10,8)+(0,5)$);
            \draw[thick, red, opacity=0.3]
              (bottomRight) -- ($(10,8)+(9,0)$);
        \end{tikzpicture}
    }
    \caption{Schematic overview of related SciML approaches and their approximate use of domain knowledge and data. At the two extremes, supervised ML learns on labeled data as input while physics-based models build up from first principles, fully encoding the underlying physics of a system. Our proposed approach, labeled as dynamics-encoded deep learning (DEDL), requires relatively low amounts of data and incorporates domain knowledge differently depending on the problem of dynamics discovery or parameter estimation.}
    \label{fig:survey}
\end{figure}

A key challenge in SciML lies in choosing how to leverage and effectively integrate domain knowledge into a computational framework. In \cref{fig:survey}, we compare some well-known ``physics-based" SciML approaches in the literature; see \cite{PeNNsurvey2024, Kim2021}. Some approaches ensure alignment with the underlying known physics by using data originating from a physical system, as in physics-guided neural networks (PgNNs) \cite{daw2021PGNN}. Other approaches apply regularization with domain knowledge, as in physics-informed neural networks (PINNs) \cite{Raissi2019PINN}, whereas physics-encoded neural networks (PeNNs) \cite{Rao2021PeNN} build these laws directly into the model architecture or representation. Furthermore, neural ODEs \cite{NODE} learn the underlying continuous-time dynamics or and neural operators \cite{Lu2021DeepONet} learn mappings between entire functions. 
Most existing works in the SciML literature assume that the observed data are either contaminated with weak Gaussian noise or no noise at all. For example, several approaches assume Gaussian noise in the context of sparse and irregular data \cite{Chen2021SRPINN, Goyal2023NODE} or uncertainty quantification \cite{Yang2021BPINN} while others consider non-Gaussian noise \cite{Pilar2024nonGaussian}. 
Within this landscape of SciML methods, we present two approaches for incorporating domain knowledge into learning aspects of dynamical systems characterized by noisy data.

\paragraph{Problem Overview}
We consider dynamical systems of the form 
\begin{equation}
    \frac{d\vec{x}}{dt} = \vec{f}\left(t,\vec{x};\lambda\right), \quad \vec{x}(t_0) = \vec{x}_0
    \label{eq:ODE}
\end{equation}
where $\Vec{x} = \Vec{x}(t) \in \R^n$ is a vector representing the states of the system at time $t$, the RHS function $\Vec{f}:\R \times \R^n \times \R^{|\lambda|} \to \R^n$ is a mapping describing the system dynamics, and $\Vec{\lambda}\in \R^{|\lambda|}$ is a vector of the system physical parameters. In this work specifically, $\Vec{\lambda}$ represents constants parameterizing the RHS mapping $\Vec{f}$. Here, $|\lambda|$ denotes the number of physical parameters. We assume that $t_0 = 0$, and $\vec{x}_0 \in \R^n$ is a given initial condition.
Given corrupt observations of the system states $\Vec{x}(t)$ at some discrete times,
\begin{equation}
    \Vec{y}_j = \Vec{x}(t_j) + \varepsilon_j, \quad \varepsilon_j\sim\mathcal{N}(\Vec{0},\Gamma_\text{obs}), \quad j = 1, 2, \dots,
\end{equation}
we aim to learn the relevant missing physics: either the unknown dynamics $\vec{f}$ and/or unknown physical parameters $\Vec{\lambda}$. In real-world applications of these problems, the observed data are typically corrupted with significant levels of noise and limited, being infrequent in terms of time or space.

System identification aims to discover the unknown dynamics $\vec{f}$ of a system modeled as in \eqref{eq:ODE} using only observed data from that system. These problems can result in {equation discovery} by finding an approximate representation of the governing equations \cite{Brunton2016SINDy, Chen2021SRPINN} or {dynamics discovery} by determining an approximation of the dynamics \cite{Raissi2018HPM, Churchill2023Missing, NODE}. Equation discovery requires prior knowledge of candidate functions to learn the unknown weights of each of these terms and build the governing equations which is feasible in some applications where the physics or underlying scientific knowledge is well studied. However, in this work, we do not assume that we have an understanding of candidate functions that could be included in the governing equations and, instead, learn the system dynamics solely from data. In fact, we do not discover the governing equations explicitly but implicitly determine an approximation of the dynamical system for prediction.

Parameter estimation, on the other hand, aims to estimate the physical parameters $\vec{\lambda}$ of the system in \eqref{eq:ODE} given the available system observations as well as knowledge of the forward model $\vec{f}$. There are a variety of deterministic and statistical methods in the literature used to estimate system parameters when the governing equations are known; see, e.g., \cite{Kaipio2005ParamEst, viana2021survey, Tarantola2005IP, Johnson1992, Banks2014, DRAM2006, LiuWest2001, Arnold2013, Evensen2009, Arnold2014, Arnold2023}. In the deep learning setting, some approaches use a neural network approximation for the physical parameters \cite{Tipireddy2022LMM,Gaskin2023Param} while others add the physical parameters to the neural network to be optimized along with the weights and biases \cite{Raissi2019PINN, Grimm2022SIR}. 
For the parameter estimation problem, our goal is to approximate the states of the dynamical system in addition to the unknown physical parameters.

\paragraph{Paper Contributions and Organization}
In this work, we propose a general approach for incorporating domain knowledge into neural network training for solving missing physics problems given noisy system observations. 
In particular, we develop a robust computational framework that considers the case of uniformly observed time series data corrupted with significant Gaussian noise. 
Depending on the problem at hand, we encode underlying physics information relating to the system in different ways and at different points of the training process to effectively learn the unknown dynamics or physical parameter values and make accurate model predictions of the dynamical system states.
Our main contributions are as follows:
\begin{itemize}
 \item For dynamics discovery, we combine deep neural networks and traditional numerical methods for solving differential equations in a novel approach for learning completely unknown system dynamics. In particular, we use a neural network to approximate the unknown dynamics and train the neural network with a numerical approximation of the states and observed trajectories from the system.
 \item To address the parameter estimation inverse problem, we encode the underlying physics by generating a neural network approximation of the states at discretized time steps and training the neural network in two phases. We use a numerical scheme with randomly initialized physical parameter values to pre-train the neural network. We then fine-tune the neural network to learn the unknown physical parameters using observed data from the system. Computational analysis highlights the importance of pre-training in improving the model predictions by properly initializing its parameters.
\end{itemize}

This framework effectively combines deep learning and classic numerical methods to address dynamics discovery and parameter estimation inverse problems given corrupt system observations.
The remainder of the paper is organized as follows. \Cref{sec:missing,,sec:inverse} detail the approaches for dynamics discovery and parameter estimation, respectively, in this framework and apply these methods to a suite of nonlinear test problems that exhibit oscillatory and chaotic dynamics. 
\Cref{sec:discussion} provides a discussion of results and concludes the paper with future considerations.

\section{System Identification via Dynamics Discovery}
\label{sec:missing}
\subsection{Learning Completely Unknown Dynamics}
\label{sec:missing-method}
\begin{figure}[t!]
    \centering
    \includegraphics[width=0.5\textwidth]{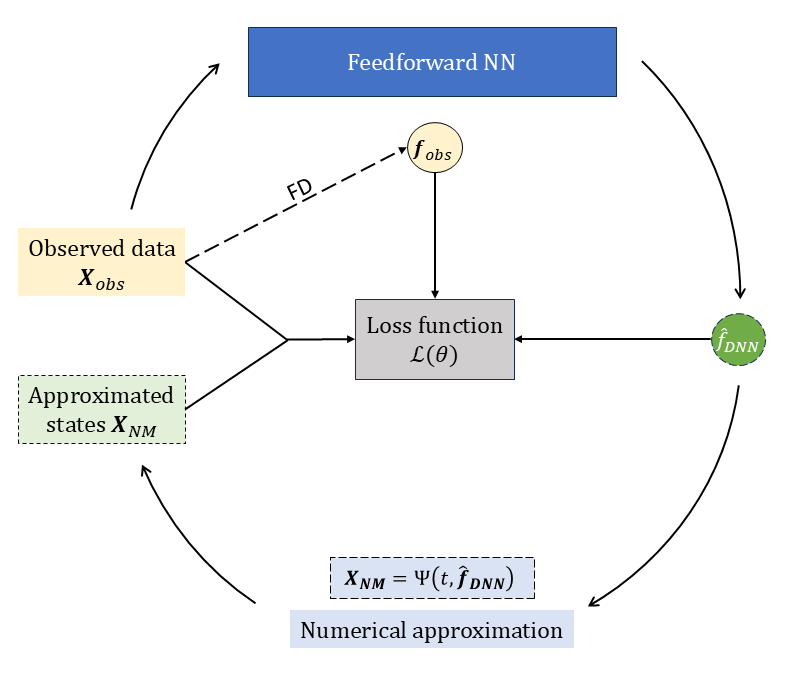}
    \caption[Schematic for proposed neural network to model dynamical systems with completely unknown dynamics]{Schematic for the proposed neural network with numerical ODE methods to model dynamical systems with completely unknown dynamics. We obtain an approximation for the RHS dynamics $\hat{\vec{f}}_{\textit{DNN}}$ using the observed states as input, and then use $\hat{\vec{f}}_{\textit{DNN}}$ to compute a numerical approximation of the states $\Vec{X}_{\textit{NM}}$. The observed states and model predictions are used in the loss function to update $\hat{\vec{f}}_{\textit{DNN}}$. The approximation $\Vec{f}_{obs}$ encodes information about the RHS dynamics (obtained via finite difference from the system observations) into the dynamics loss term.
    }
    \label{fig:missing-method}
\end{figure}

The goal of dynamics discovery is to determine an approximation for the system dynamics that can be used for prediction. That is, we approximate the unknown RHS dynamics $\vec{f}$ with a vector $\vec{\hat{f}} \in \R^n$ at some discretized time steps $t \in \R$, given the observed data from the system, denoted as $\vec{X}_{obs}$. We learn the unknown dynamics $\vec{f}$ using a feed-forward neural network with the observed states as input. We then compute a numerical approximation of the states $\vec{X}_{\textit{NM}}$ using the neural network output $\hat{\vec{f}}_{\textit{DNN}}$ and utilize the observed and approximated states in the loss function to update the RHS dynamics $\hat{\vec{f}}_{\textit{DNN}}$. To train the neural network, we consider the following loss function:
\begin{equation}
    \mathcal{L}(\theta) = \mathcal{L}_{ic}(\theta) + \mathcal{L}_{dyn}(\theta) + \mathcal{L}_{obs}(\theta)
    \label{eq:missing-loss}
\end{equation}
where
\begin{align}
    \mathcal{L}_{ic}(\theta) &= \frac{1}{N_{ic}}\norm{\Vec{X}_{\textit{NM}}(t_0; \theta)-\Vec{X}_0(t_0)}_2^2 \label{eq:missing-ic-loss}\\
    \mathcal{L}_{dyn}(\theta) &= \frac{1}{N_{dyn}}\norm{\hat{\vec{f}}_{\textit{DNN}}(t;\theta)- \Vec{f}_{obs}(t)}_2^2 \label{eq:missing-dynamics-loss}\\
    \mathcal{L}_{obs}(\theta) &= \frac{1}{N_{obs}}\norm{\Vec{X}_{\textit{NM}}(t;\theta) - \Vec{X}_{obs}(t)}_2^2  \label{eq:missing-obs-loss}
\end{align}
and the vector $\theta$ contains the neural network weights and biases. Here, $\norm{\cdot}_2^2$ denotes the square of the $\ell^2$-norm. All the above terms are written in vector notation, so $\Vec{X}_0$ represents a vector of the initial conditions, and $\Vec{X}_{\textit{NM}}$ denotes the approximate solution given from the numerical schemes. \Cref{eq:missing-ic-loss} is the initial condition loss comparing the given initial conditions from the problem with the numerical approximation of the states at $t = 0$. The dynamics loss term in \cref{eq:missing-dynamics-loss} compares the neural network output $\hat{\vec{f}}_{\textit{DNN}}$ with $\vec{f}_{obs}$, which is computed from $\vec{X}_{obs}$ using an appropriate derivative approximation technique.
The observation loss term in \cref{eq:missing-obs-loss} accounts for differences in the approximated states $\Vec{X}_{\textit{NM}}$ and the observed states $\vec{X}_{obs}$. 
During each epoch of training, the learned RHS dynamics $\hat{\vec{f}}_{\textit{DNN}}$ updates with adjustments made through comparisons with $\Vec{f}_{obs}$ in the loss function. We use the neural network output to compute $\Vec{X}_{\textit{NM}}$ as follows:
\begin{equation}
    \vec{X}_{\textit{NM}} = \Psi(t,\hat{\vec{f}}_{\textit{DNN}})
\end{equation}
where $\Psi(t, \cdot)$ denotes a general numerical scheme used to approximate the solution $\vec{X}_{\textit{NM}}$ at time steps $t$. We utilize Runge-Kutta (RK) and linear multistep methods (LMMs) as our numerical ODE solvers for the examples in \cref{sec:missing-results,,sec:inverse-results}. We consider both explicit and implicit methods to account for various types of systems that may be stiff and complex. For clarity, the numerical approximation $\vec{X}_{\textit{NM}}$ will be hereafter referred to as the model prediction in the dynamics discovery problem since it uses the neural network approximation of the RHS to compute the state variables. \Cref{fig:missing-method} illustrates the interaction between the neural network output with the numerical method for dynamics discovery, and \cref{alg:missing} provides pseudocode for the neural network training.

\begin{algorithm}[t!]
    \caption{Neural network training for completely unknown dynamics} \label{alg:missing}
    \begin{algorithmic}
        \REQUIRE Observed values from system $\Vec{X}_{obs}$, initial conditions $\Vec{X}_0$
        \ENSURE RHS approximation $\hat{\vec{f}}_{\textit{DNN}}$ at time steps $t$
        \STATE Define an optimizer \textit{optim}
        \WHILE{\textit{optim} tolerance not met}
        \STATE Reset gradients of \textit{optim}
        \STATE $\hat{\vec{f}}_{\textit{DNN}} = \textit{DNN}(\Vec{X}_{obs})$
        \STATE Compute required gradients of $\Vec{X}_{obs}$
        \STATE Compute $\vec{X}_{\textit{NM}} = \Psi(t,\hat{\vec{f}}_{\textit{DNN}})$
        \STATE Compute initial condition loss term: $\mathcal{L}_{ic}(\theta) = \frac{1}{N_{ic}}\norm{\Vec{X}_{\textit{NM}}(t_0; \theta)-\Vec{X}_0(t_0)}_2^2$ 
        \STATE Compute dynamics loss term: $\mathcal{L}_{dyn}(\theta) = \frac{1}{N_{dyn}}\norm{\hat{\vec{f}}_{\textit{DNN}}(t;\theta)- \Vec{f}_{obs}(t)}_2^2$
        \STATE Compute observation loss term: $\mathcal{L}_{obs}(\theta) = \frac{1}{N_{obs}}\norm{\Vec{X}_{\textit{NM}}(t;\theta) - \Vec{X}_{obs}(t)}_2^2$ 
        \STATE $\mathcal{L}(\theta) = \mathcal{L}_{ic}(\theta) + \mathcal{L}_{dyn}(\theta) + \mathcal{L}_{obs}(\theta)$ 
        \STATE Compute gradients of $\mathcal{L}(\theta)$ using backpropagation
        \STATE Update $\theta$ by taking an \textit{optim} step
        \ENDWHILE
    \end{algorithmic}
\end{algorithm}

\subsection{Numerical Experiments for Dynamics Discovery}
\label{sec:missing-results}
We apply the method described in \cref{sec:missing-method} to a suite of test problems with oscillatory and chaotic dynamics.
\subsubsection{General Setup}
\label{sec:results-methodology}
To generate noisy simulated data for our numerical examples, we assume that the observed states are corrupted by Gaussian noise with mean zero and covariance matrix prescribed based on a percentage of the standard deviations of the exact solution trajectories. More specifically, for some discrete times $t_j$, $j=1,\dots,T$, we let
\begin{equation}
\vec{X}_{obs}(t_j) = \vec{X}_{exact}(t_j) + \delta \cdot \eta_j, \quad \eta_j\sim\mathcal{N}(\vec{0},\mathsf{\Gamma})
\end{equation}
where $\delta$ represents a user-defined noise level, $\mathsf{\Gamma}$ is an $n\times n$ diagonal matrix with the variance of each state along the main diagonal, and $\mathcal{N}(\cdot,\cdot)$ denotes the multivariate normal distribution.

We compare various numerical schemes for computing the state approximations in each method, employing both RK and LMMs within the neural network. In particular, we utilize the explicit Runge–Kutta–Fehlberg method of order 4(5), denoted RK45, for computational efficiency and consider methods from the three main LMM families: Adams-Bashforth (AB), Adams-Moulton (AM), and the backwards differentiation formulae (BDF). In this work, we denote a specific LMM scheme using its family name and number of steps. For example, AB2 refers to the 2-step Adams-Bashforth method. The AB methods are explicit, but both the AM and BDF families are implicit and require the solution of an additional optimization problem at each time step. 
When choosing a numerical ODE solver, the size and shape of the absolute stability regions are critical because they determine which values of the time step will give bounded solutions. In this section, we show results for a suitable choice of numerical method (i.e., one that is stable and well-suited for the problem at hand). 
We note that 2-step methods provide a reasonable balance between computational efficiency and accurate results for the examples considered in this work.

We use Python3 and PyTorch \cite{pytorch} to construct and train the neural network architectures described in this section. 
Results were produced locally using an Acer Aspire A515-57 laptop computer with 16 GB RAM and an Intel\textsuperscript{\textregistered} Core\textsuperscript{\texttrademark} i7-1255U processor.
For neural network training, we utilize feed-forward neural networks with varying sizes based on the complexity of the problem. Each layer besides the output layer uses $\tanh(x)$ as the activation function. We use Adam \cite{adam} followed by L-BFGS \cite{LBFGS} as the optimization algorithm for computational efficiency. 
More specifically, we apply an Adam optimizer with an initial learning rate of 0.001, followed by the L-BFGS optimizer with a learning rate of 1.0 and a maximum number of iterations of 50,000.
To compute $\Vec{f}_{obs}$ from the available data, we employ finite difference approximation. We note that while a finite difference approach was sufficient for the time steps chosen in examples presented in this work, for other problems, one may consider alternative derivative approximation techniques such as the Savitzky-Golay filter \cite{sgfilter}.

To evaluate the performance of our models, we randomly select time steps within the domain to generate a set of test observed data that the neural network was not trained on. We pass this test data into the neural network to evaluate the quality of the model predictions. 
As an evaluation metric, we compute the mean squared error (MSE) between the exact solution and neural network approximation for each state at the test data points: 
\begin{equation}
    \text{MSE} = \frac{1}{N}\sum_{i=1}^N \left(y_i - \hat{y}_i\right)^2
\end{equation}
where $y_i$ is the true value of the $i$-th sample, $\hat{y}_i$ is the corresponding predicted value, and $N$ is the number of samples in the test data set. For all the examples in this section, since we are using simulated data with known underlying true dynamics and states, we compute the MSE and relative errors using the exact solutions and parameter values compared with our model predictions and approximations. However, we acknowledge that in cases where the dynamics of a system are truly unknown, the evaluation of the model becomes more difficult and different metrics are required.

\subsubsection{Example 1: FitzHugh-Nagumo Model}
\label{sec:FNmodel-missing}
The FitzHugh-Nagumo model is a simplified version of the Hodgkin-Huxley model describing neural spike generation and propagation \cite{FitzHugh1961, Nagumo1962, Hodgkin1952}. The behavior is characterized by a short spike of membrane voltage, $v(t)$, which is diminished over time by a slower recovery variable, $w(t)$. The standard form of the governing equations for this dynamical system is
\begin{align}
    \frac{dv}{dt} &= v-\frac{v^3}{3} - w + z\label{eq:fn-v}\\[0.2cm]
    \frac{dw}{dt} &= \frac{1}{c}\left(v + a - bw\right) \label{eq:fn-u}
\end{align}
where $a$, $b$, and $c$ are system parameters and $z$ corresponds to an applied membrane current. We set $a = 0.7$, $b = 0.8$, $c = 12.5$, and $z = 1$ as the true parameter values and take $v(0) = -2.8$ and $w(0) = -1.8$ as the initial conditions. We consider the solution obtained by using RK45 with a time step of $\Delta t = 0.0001$ as the exact solution for this example.

\begin{figure}[t!]
    \centerline{\includegraphics[width=0.33\textwidth]{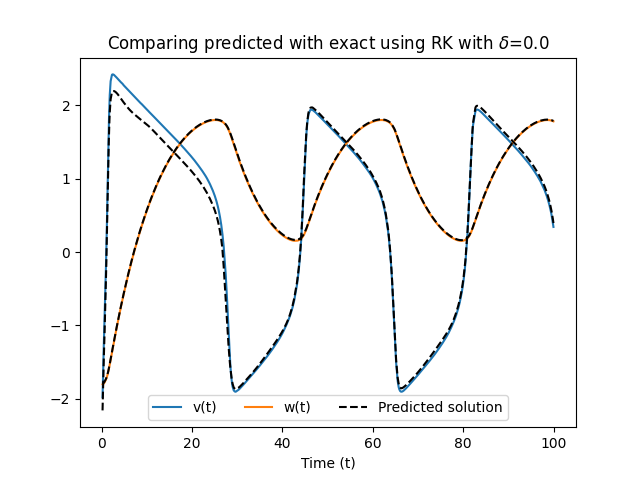} \includegraphics[width=0.33\textwidth]{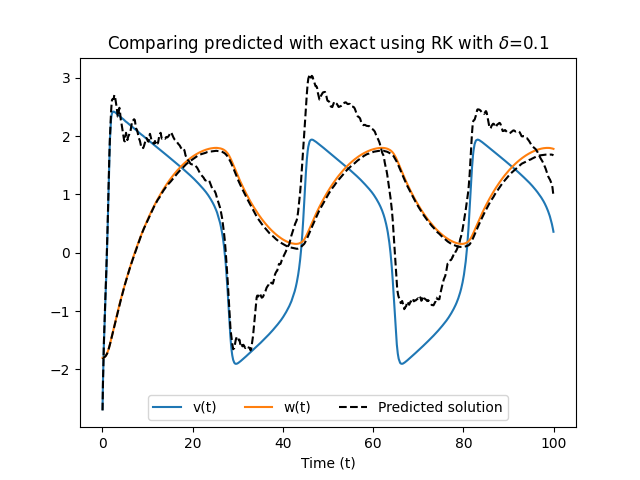} \includegraphics[width=0.33\textwidth]{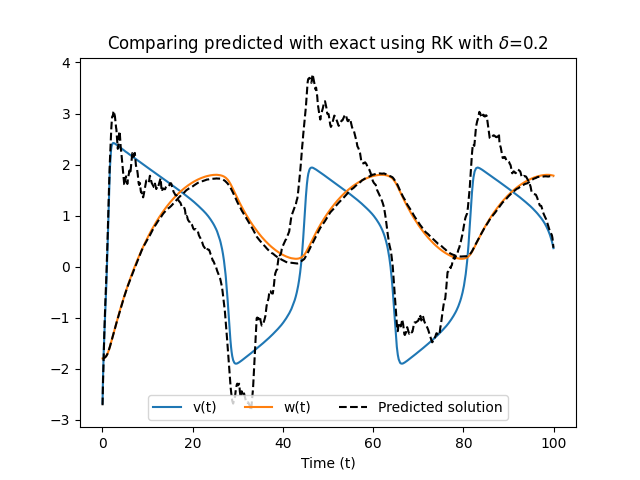}}
    \caption{Dynamics discovery model predictions of the FitzHugh-Nagumo model obtained using observed data at various noise levels ($\delta = 0, 0.1, 0.2$) with RK45.}
    \label{fig:FN-missing}
\end{figure}

\paragraph{Results}
We implement a feed-forward neural network with 4 hidden layers and 64 nodes per layer. For both the RK45 and BDF2 schemes, we use the same time step of $\Delta t = 0.1$. \Cref{fig:FN-missing} shows the resulting FitzHugh-Nagumo model predictions obtained from the neural network trained on observed data with various noise levels and using the RK scheme, and \cref{tab:FN-missing} displays the corresponding MSEs between the predictions and exact solutions.

In the case of noiseless data, we see that the model predictions of both $v(t)$ and $w(t)$ very closely match the true states for both numerical schemes. When the noise level increases, more error accrues in the predictions for $v(t)$, whereas the predictions for $w(t)$ remain consistently close to the ground truth. We attribute these results to the larger magnitude of change in the $v(t)$ signal, which corresponds to more noise in the corrupted observations. However, even though the model predictions for $v(t)$ appear more erratic, they remain consistent between the RK and LMM schemes. In fact, \cref{tab:FN-missing} indicates that the MSEs comparing the model prediction with the exact solution are on the same order of magnitude for both the RK and LMM schemes for each system component and noise level. 

\begin{figure}
    \centering
    \includegraphics[width=0.4\textwidth]{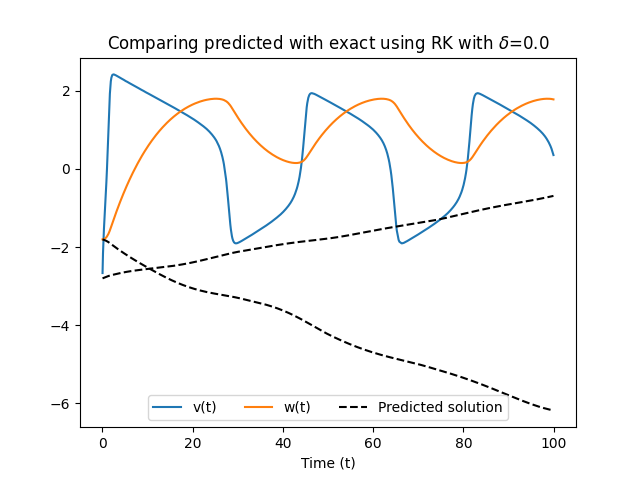} 
    \caption{Dynamics discovery model predictions of the FitzHugh-Nagumo model obtained by excluding $\mathcal{L}_{dyn}(\theta)$ from \cref{eq:missing-loss}. The MSE for $v(t)$ and $w(t)$ were 7.57e+00 and 2.79e+01, respectively.}
    \label{fig:missing-weight0}
\end{figure}

Additional experiments in \cref{app:weight} demonstrate the necessity of both an observation and dynamics loss term in this problem. In particular, the importance of the dynamics loss term is most evident when we completely bias toward $\mathcal{L}_{obs}$. As seen in \cref{fig:missing-weight0}, without the dynamics loss term \cref{eq:missing-dynamics-loss}, the model predictions indicate that the neural network did not learn the dynamics at all, and the MSE for the model predictions is order of magnitudes higher than the other values in \cref{tab:FN-missing}.

\begin{table}[t!]
    \centering \footnotesize
    \begin{tabular}{|c|ll|ll|}
    \hline
     \multirow{2}{*}{Noise Level} & \multicolumn{2}{c|}{RK} & \multicolumn{2}{c|}{BDF2} \\ \cline{2-5} 
     & \multicolumn{1}{c}{$v$} & \multicolumn{1}{c|}{$w$} & \multicolumn{1}{c}{$v$} & \multicolumn{1}{c|}{$w$} \\ \hline
     0\% & 1.34e-02 & 8.28e-05 & 1.02e-02 & 7.00e-05 \\
     10\% & 5.02e-01 & 5.05e-04 & 3.02e-01 & 3.20e-04 \\
     20\% & 7.50e-01 & 3.95e-03 & 1.66e-01 & 4.47e-03 \\ \hline
    \end{tabular}
    \caption{Comparing RK and LMM scheme MSEs for FitzHugh-Nagumo dynamics discovery model predictions.}
    \label{tab:FN-missing}
\end{table}

\subsubsection{Example 2: Lorenz-63 System}
\label{sec:lorenz-missing}
The Lorenz-63 system provides a model of atmospheric convection that is notable for its chaotic solutions for certain choices of initial conditions and parameters \cite{lorenz1963}. The governing equations are of the form 
\begin{align}
    \frac{dx}{dt} &= \sigma(y-x) \label{eq:lorenz-x}\\[0.2cm]
    \frac{dy}{dt} &= x(\rho - z)-y \label{eq:lorenz-y}\\[0.2cm]
    \frac{dz}{dt} &= xy-\beta z \label{eq:lorenz-z}
\end{align}
where the parameters $\sigma = 10$, $\beta = 8/3$, and $\rho = 28$ are chosen such that the system exhibits chaotic behavior. In addition to these parameter values, we set the initial conditions $x(0) = -8$, $y(0) = 7$, and $z(0) = 27$ to apply our method in \cref{sec:missing-method} for this example. We approximate the exact solution of the Lorenz-63 system using RK45 at a fine time step of $\Delta t = 0.00025$.

\paragraph{Results}
As in \cref{sec:FNmodel-missing}, we apply the method of discovering unknown system dynamics using observed data with noise levels of 0\%, 10\%, and 20\%. We include results using noiseless observed data to distinguish predictions due to chaos versus results due to noise in the training data. When training on noiseless data, we implement a feed-forward neural network with 4 hidden layers and 64 nodes per layer with a skip connection, which bypasses the hidden layers and adds back an identity function to the output node \cite{He2015ResNet}. This provides an alternate path for the gradient in backpropagation and reduces model overfitting, particularly for this chaotic problem. To reduce overfitting on substantially noisy data, we choose a less complex neural network architecture of 32 hidden nodes in 1 layer with a skip connection when training on noisy data. For both the RK45 and BDF2 schemes, we use a time step of $\Delta t = 0.001$. 

\begin{figure}
    \centering
    \includegraphics[width=0.7\textwidth]{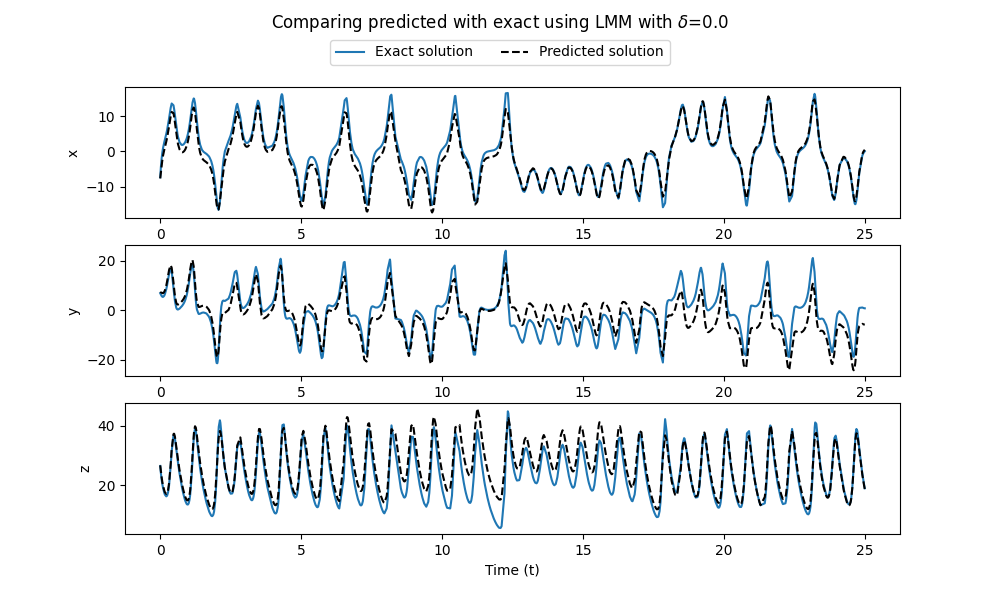}
    \includegraphics[width=0.7\textwidth]{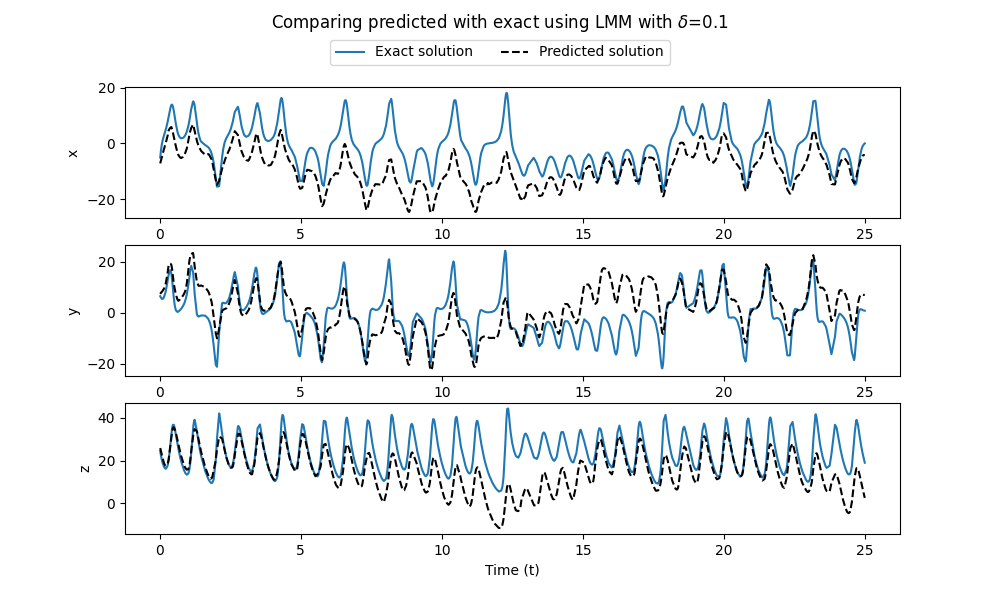}
    \includegraphics[width=0.7\textwidth]{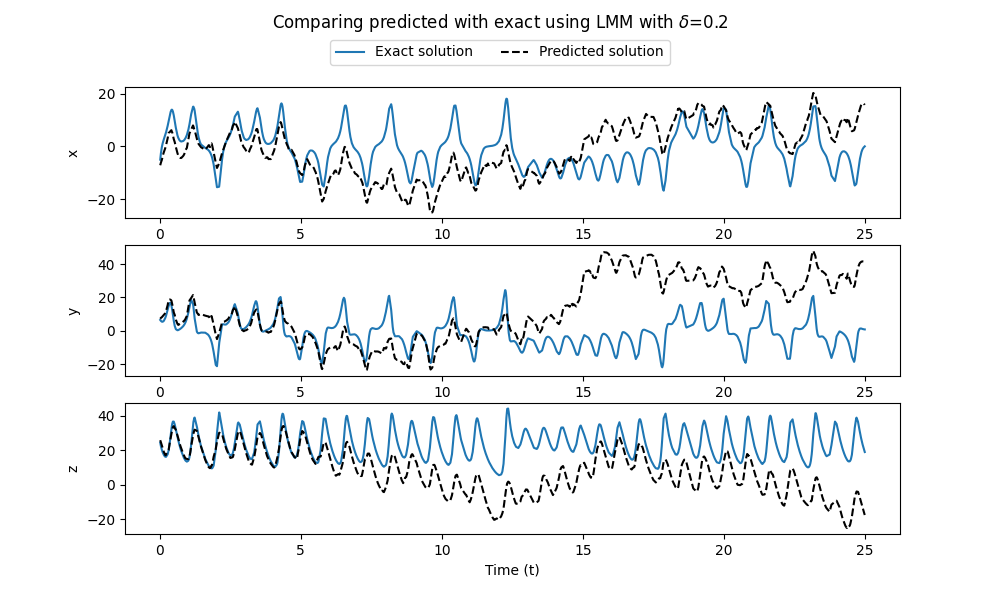}
    \caption{Dynamics discovery model predictions of the Lorenz-63 system obtained using noisy observed data at various noise levels ($\delta = 0, 0.1, 0.2$) with BDF2.}
    \label{fig:lorenz-missing}
\end{figure}

\Cref{fig:lorenz-missing} displays the predicted states of the Lorenz-63 system using the BDF2 scheme, and \cref{tab:lorenz} presents the corresponding MSEs for both RK45 and BDF2. For the noiseless case, the dynamics of the model components are fairly well recovered, specifically for $x(t)$; however, the predictions for $y(t)$ and $z(t)$ include some errors in tracking the amplitude of the trajectories from the exact solution. 
As the noise level increases in \cref{fig:lorenz-missing}, we observe state predictions drifting away from the exact solutions, particularly in the $y(t)$ and $z(t)$ components. This ``drifting" results from error propagation within the neural network, but in the $z(t)$ component, the state prediction appears to recover some of the dynamics at $t = 20$ after drifting away for $t > 8$. However, similar to \cref{sec:FNmodel-missing}, the MSEs in \cref{tab:lorenz} are again on the same order of magnitude for both the RK and LMM schemes for each system component and noise level, indicating reasonable predictions in the presence of corrupted data. 

\begin{table}[t!]
    \centering \footnotesize
    \begin{tabular}{|c|lll|lll|}
    \hline
    \multirow{2}{*}{Noise Level} & \multicolumn{3}{c|}{RK} & \multicolumn{3}{c|}{BDF2} \\ \cline{2-7} 
     & \multicolumn{1}{c}{$x$} & \multicolumn{1}{c}{$y$} & \multicolumn{1}{c|}{$z$} & \multicolumn{1}{c}{$x$} & \multicolumn{1}{c}{$y$} & \multicolumn{1}{c|}{$z$} \\ \hline
     0\% & 2.41e+00 & 2.33e+01 & 1.05e+01 & 2.83e+00 & 2.74e+01 & 1.47e+01 \\
     10\% & 9.35e+01 & 8.75e+01 & 1.07e+02 & 8.24e+01 & 7.22e+01 & 1.24e+02 \\
     20\% & 1.08e+02 & 6.04e+02 & 4.10e+02 & 9.35e+01 & 5.88e+02 & 3.99e+02 \\ \hline
    \end{tabular}
    \caption{Comparing RK and LMM scheme MSEs for Lorenz-63 dynamics discovery model predictions.}
    \label{tab:lorenz}
\end{table}

\subsubsection{Example 3: 1D Heat Equation}
\label{sec:heat-missing}
The heat equation models how heat diffuses through a spatial region as a PDE; see \cite{Holmes2019}. We consider the one-dimensional (1D) heat equation, modeling the change in temperature over time as
\begin{equation}
    \fpar{u}{t} = k \frac{\partial^2 u}{\partial x^2}, \quad 0 \leq x \leq L,\quad t \geq 0
    \label{eq:heat}
\end{equation}
with constant diffusion coefficient $k$ and the following initial and Robin boundary conditions:
\begin{align}
    u(x,0) &= \sin\left(\frac{\pi x}{2}\right),\\
    u(0,t) &= 0,\\
    \frac{\partial u}{\partial x}(L,t) &= 0 \label{eq:robin_BC2}.
\end{align}
The analytic solution to this problem using separation of variables and Fourier series is 
\begin{equation}
    u(x,t) =  \sin\left(\frac{\pi x}{2}\right)\exp\left(-\frac{\pi^2t}{4}\right).
    \label{eq:heat-exact}
\end{equation}
For this example, we assume that $k = 1$, $L = 1$, and $ t \in [0,2.5]$. 

\paragraph{Method of Lines Approximation}
\label{sec:heat-eq-mol}
The {method of lines} (MOL) approach for numerically solving PDEs discretizes the spatial dimension first, resulting in a system of coupled ODEs where each component corresponds to the solution at some spatial grid point as a function of time \cite{MOL1975}. We then compute the approximate solution of the ODE system using a numerical method such as RK or LMM. Discretizing the spatial variable into $M+1$ points, we replace the spatial derivative in \cref{eq:heat} using the second-order central difference formula:
\begin{equation}
    \frac{\partial^2 u}{\partial x^2} \approx D_0^2u = \frac{u_{i+1}-2u_i + u_{i-1}}{h^2}
\end{equation}
where $u_i = u_i(t) \approx u(x_i,t)$ and $h = \Delta x$ denotes the spatial step size, with approximation error $\mathcal{O}(h^2)$. This discretization leads to the following MOL approximation of \cref{eq:heat}:
\begin{equation}
    \frac{du_i}{dt} = k\frac{u_{i+1}-2u_i + u_{i-1}}{h^2}, \quad 1 \leq i \leq M-1.
    \label{eq:heat-MOL}
\end{equation}
Spatial discretization of the initial and boundary conditions yields
\begin{align}
    u_{i}(0) &= \sin\left(\frac{\pi x_i}{2}\right), \quad 0 \leq i \leq M, \label{eq:heat-eq-ic}\\
    u_0(t) &= 0, \quad t \geq 0,\\
    \frac{u_{M}(t) - u_{M-1}(t)}{h} &= 0, \quad t \geq 0,
\end{align}
using a forward difference approximation for \cref{eq:robin_BC2}, which implies that $u_{M}(t) = u_{M-1}(t)$.

\begin{figure}[t!]
    \centerline{\includegraphics[width=0.5\textwidth]{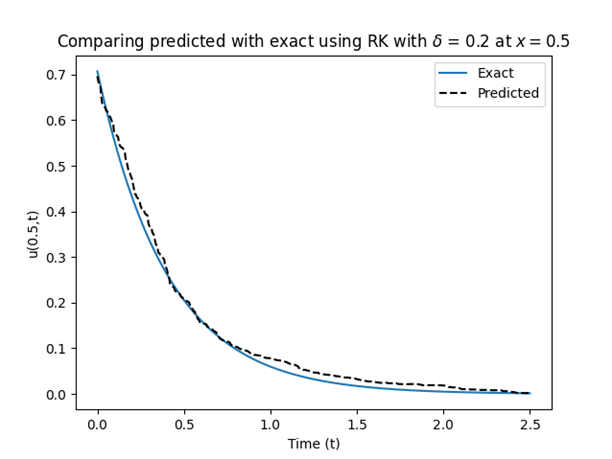} \includegraphics[width=0.5\textwidth]{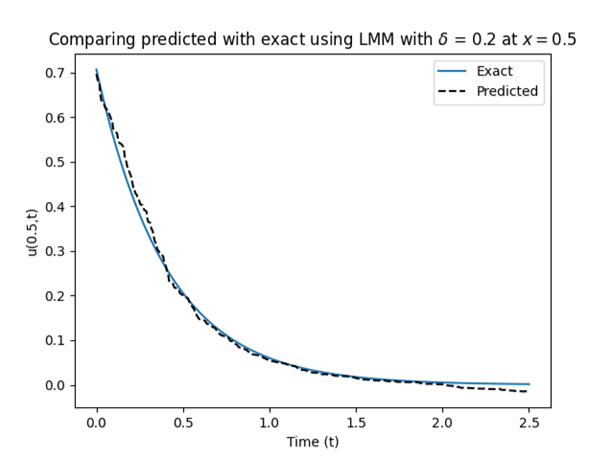}}
    \caption{Dynamics discovery model predictions of the heat equation at spatial location $x=0.5$ obtained using 20\% noisy data with RK45 and BDF2.}
    \label{fig:heat-missing-noise}
\end{figure}

\begin{figure}[t!]
    \centering
    \includegraphics[width=\textwidth]{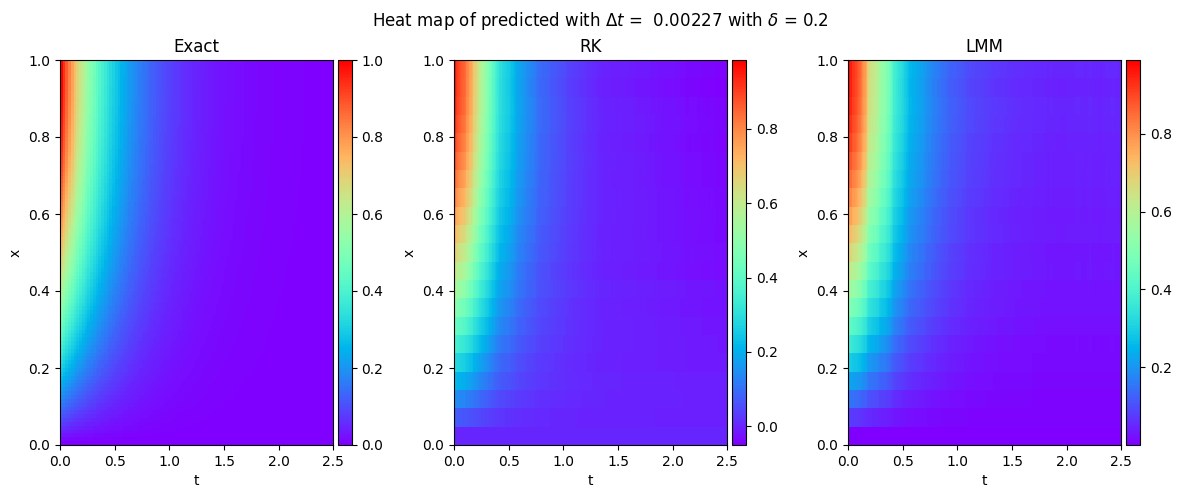}
    \caption{Heat map of dynamics discovery model predictions for the heat equation with 20\% noisy data.}
    \label{fig:heat-missing-heatmap}
\end{figure}

\paragraph{Results}
We implement a neural network architecture with the 64 hidden nodes per layer and 2 hidden layers. For both numerical schemes, we discretize the spatial variable into $M+1$ equispaced points where $M=20$ and use a time step of $\Delta t = 0.00227$ to account for stability of the numerical methods.
\Cref{fig:heat-missing-noise} shows the heat equation model predictions at spatial location $x=0.5$ with 20\% noisy data for both RK45 and BDF2 schemes. These results are similar to the previous two examples, given appropriate choices for the spatial and temporal discretizations. We observe close predictions to the ground truth at $x=0.5$ and correspondingly small MSEs in \cref{tab:heat}. While not explicitly shown, model predictions at different spatial locations are similar to those at $x=0.5$ as seen in \cref{fig:heat-missing-heatmap}. 

\begin{table}[t!]
    \centering\footnotesize
    \begin{tabular}{|c|c|c|}
    \hline
      Noise Level & RK & BDF2 \\ \hline
      10\% & 3.03e-03 & 1.74e-03 \\
     20\% & 2.86e-03 & 2.78e-03 \\ \hline
    \end{tabular}
    \caption{Comparing RK and LMM scheme MSE for the heat equation dynamics discovery model predictions.}
    \label{tab:heat}
\end{table}

\section{Parameter Estimation}
\label{sec:inverse}
\subsection{Learning Unknown Physical Parameters}
\label{sec:inverse-method}
\begin{figure}[t!]
    \centering
    \includegraphics[width=0.9\textwidth]{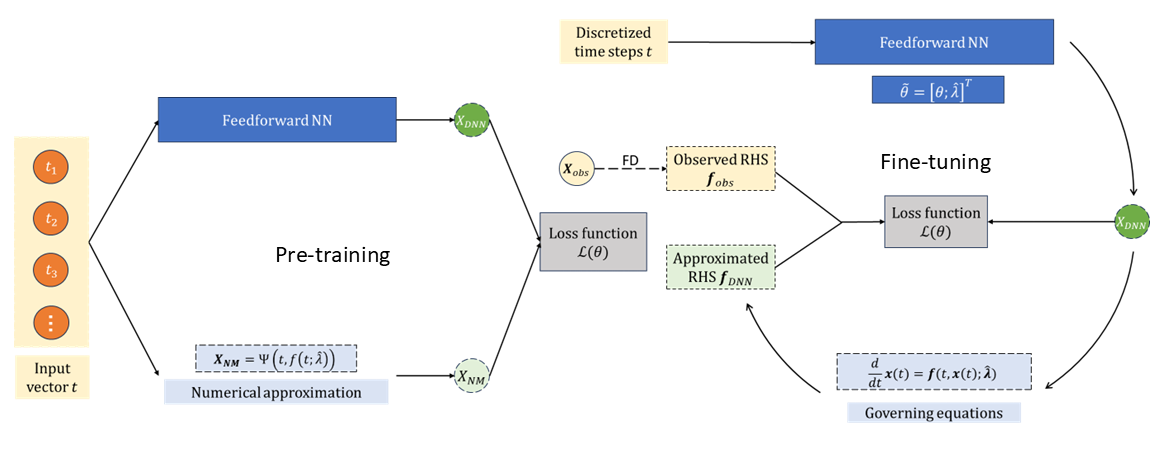}
    \caption[Schematic for proposed neural network to model dynamical systems with completely unknown physical parameters]{Schematic for the proposed neural network with numerical ODE methods to model dynamical systems with completely unknown physical parameters. During the pre-training phase, we use the neural network output and numerical approximation computed using $\vec{\hat{\lambda}}$ to initialize the neural network parameters $\Vec{\theta}$ and encode the known governing equations. Then, in fine-tuning, we append the pre-trained network parameters $\theta$ with $\vec{\hat{\lambda}}$ to optimize the unknown physical parameters along with the network parameters during backpropagation. }
    \label{fig:inverse-method}
\end{figure}

For parameter estimation, we aim to determine the value of unknown physical parameters given a complete functional form of the governing equations and observed data from the system. We generate a neural network approximation of the system states at discretized time steps and train the neural network to learn the unknown physical parameters using the governing equations that describe the RHS dynamics. However, unlike the forward prediction problem considered in \cite{Raissi2019PINN}, here, the RHS vector includes error from the estimated physical parameter values $\vec{\hat{\lambda}} \in \R^{|\lambda|}$. We begin by pre-training the neural network with randomly initialized values for the physical parameters and then fine-tune the physical parameter approximations along with the neural network parameters, as illustrated in \cref{fig:inverse-method}. Since we assume the physical parameter values are unknown and the observed data are significantly noisy, pre-training the neural network parameters provides a better starting point for the neural network that is then refined during fine-tuning to perform the physical parameter estimation.

We use the following general loss function in both phases of neural network training:
\begin{equation}
    \mathcal{L}(\theta) = \mathcal{L}_{ic}(\theta) + \mathcal{L}_{dyn}(\theta) + \mathcal{L}_{obs}(\theta) + \sum_{i=1}^{|\lambda|} \mathcal{L}_{\lambda_i}
    \label{eq:inv-loss}
\end{equation}
where $\mathcal{L}_{ic}(\theta)$, $\mathcal{L}_{dyn}(\theta)$, and $\mathcal{L}_{obs}(\theta)$ represent the initial condition, dynamics, and observation loss terms, respectively, and each $\mathcal{L}_{\lambda_i}$ is a parameter loss term enforcing a range of values for each system parameter $\vec{\lambda}_i$, $i = 1,\dots, |\lambda|$. The work of Wang, Teng, and Perdikaris (2021) discovered a bias toward the physics loss term of a PINN due to vanishing gradients associated with the boundary loss term \cite{Wang2021PINN}. As a result, we chose our optimization algorithm very carefully since our proposed approaches similarly implement a constrained loss. We employ different training strategies for pre-training and fine-tuning. Pre-training uses an Adam optimizer with initial learning rate of 0.001, followed by L-BFGS optimizer with maximum number of iterations of 20,000. For fine-tuning, we utilize the same optimization procedure (Adam followed by L-BFGS) but with a smaller initial learning rate of 0.0001 and maximum iteration of 5,000. To account for the vanishing gradient of the initial condition loss term in the fine-tuning phase of our approach, we add a weight of $10^{3}$ in front of our initial condition loss term in \eqref{eq:inv-loss} to improve neural network training. 

\subsubsection{Pre-training Phase}
\label{sec:inv-pt}

To begin the pre-training phase, we randomly initialize the physical parameter values by drawing each from a uniform distribution with lower and upper bounds set to enforce a likely range of values for the parameter. We then compute a numerical approximation $\Vec{X}_{\textit{NM}}$ for the states at the discretized time steps using these parameter values in the known RHS, implicitly encoding the governing equations of the system, and the compare $\Vec{X}_{\textit{NM}}$ with the neural network output $\Vec{X}_{\textit{DNN}}$. This pre-training procedure prepares the neural network for estimating the physical parameters by initializing the neural network parameters $\Vec{\theta}$ such that the neural network structure implicitly incorporates the governing equations. This makes the training process more efficient and robust. In fact, as shown in \cref{sec:FNmodel-inv}, we observe a performance degradation by training the neural network directly without pre-training on the numerical approximation. We use the following terms in \cref{eq:inv-loss} for pre-training:
\begin{align}
    \mathcal{L}_{ic}(\theta) &= \frac{1}{N_{ic}}\norm{\Vec{X}_{\textit{DNN}}(t_0;\theta)-\Vec{X}_0(t_0)}_2^2\label{eq:inv-ic-loss-pt}\\
    \mathcal{L}_{dyn}(\theta) &= \frac{1}{N_{dyn}}\norm{\Vec{X}_{\textit{DNN}}(t;\theta) - \Vec{X}_{\textit{NM}}(t;\vec{\hat{\lambda}})}_2^2\label{eq:inv-dynamics-loss-pt}
\end{align}
and $\mathcal{L}_{obs}(\theta) =  \mathcal{L}_{\lambda_i} = 0$ for all $i=1,\dots,|\lambda|$. As opposed to the dynamics loss in \cref{eq:missing-dynamics-loss} from \cref{sec:missing-method}, which encodes the known dynamics given by the observed data, Equation~\cref{eq:inv-dynamics-loss-pt} in this pre-training phase encodes the known functional form of the RHS given by the numerical solution $\vec{X}_{\textit{NM}}$. \Cref{alg:inverse-pt} provides pseudocode for the neural network pre-training. With each epoch of training, the approximation for the states improve as the neural network parameters update during backpropagation. However, since this step uses randomly initialized physical parameters, the resulting approximation is far from exact, and we require fine-tuning to improve $\Vec{X}_{\textit{DNN}}$ further. 

\begin{algorithm}[t!]
    \caption{Neural network pre-training with uniform random initial physical parameters} \label{alg:inverse-pt}
    \begin{algorithmic}
        \REQUIRE Discretized time steps $t$, initial conditions $\Vec{X}_0$
        \ENSURE State approximation $\Vec{X}_{\textit{DNN}}$ at time steps $t$, pre-trained network parameters $\Vec{\theta}$
        \STATE Define an optimizer \textit{optim}
        \STATE Randomly initialize $\Vec{\hat{\lambda}}$ 
        \WHILE{\textit{optim} tolerance not met}
        \STATE Reset gradients of \textit{optim}
        \STATE $\Vec{X}_{\textit{DNN}} = \textit{DNN}(t)$
        \STATE Compute $\vec{X}_{\textit{NM}} = \Psi(t,\vec{f}(t; \vec{\hat{\lambda}}))$
        \STATE Compute initial condition loss term: $\mathcal{L}_{ic}(\theta) = \frac{1}{N_{ic}}\norm{\Vec{X}_{\textit{DNN}}(t_0;\theta)-\Vec{X}_0(t_0)}_2^2$
        \STATE Compute dynamics loss term: $\mathcal{L}_{dyn}(\theta) = \frac{1}{N_{dyn}}\norm{\Vec{X}_{\textit{DNN}}(t;\theta) - \Vec{X}_{\textit{NM}}(t;\vec{\hat{\lambda}})}_2^2$     
        \STATE $\mathcal{L}(\theta) = \mathcal{L}_{ic}(\theta) + \mathcal{L}_{dyn}(\theta)$
        \STATE Compute gradients of $\mathcal{L}(\theta)$ using backpropagation
        \STATE Update $\theta$ by taking an \textit{optim} step
        \ENDWHILE
    \end{algorithmic}
\end{algorithm}

\subsubsection{Fine-tuning Phase}
\label{sec:inv-ft}
In the fine-tuning phase of training the neural network, we use the observed data from the system to fine-tune the neural network parameters $\Vec{\theta}$ taken from the pre-trained network. We augment the physical parameters $\vec{\hat{\lambda}}$ to the neural network parameters $\Vec{\theta}$ and treat them as trainable parameters $\tilde{\theta} = (\theta, \vec{\hat{\lambda}})^T$ in the fine-tuning neural network. The physical parameters are also updated at each epoch when using backpropagation and the optimization algorithms to train the neural network. Thus, we fine-tune the neural network by updating $\tilde{\theta}$ using observed data and learning the approximated physical parameters $\Vec{\hat{\lambda}}$ in the process. From \cref{eq:inv-loss}, we use the following loss terms to fine-tune the neural network: 
\begin{align}
    \mathcal{L}_{ic}(\tilde{\theta}) &= \frac{1}{N_{ic}}\norm{\Vec{X}_{\textit{DNN}}(t_0;\tilde{\theta})-\Vec{X}_0(t_0)}_2^2\label{eq:inv-ic-loss-ft}\\
    \mathcal{L}_{dyn}(\tilde{\theta}) &= \frac{1}{N_{dyn}}\norm{\hat{\vec{f}}_{\textit{DNN}}(t; \tilde{\theta})- \Vec{f}_{param}(t; \vec{\hat{\lambda}})}_2^2\label{eq:inv-dynamics-loss-ft}\\
    \mathcal{L}_{obs}(\tilde{\theta}) &= \frac{1}{N_{obs}}\norm{\Vec{X}_{\textit{DNN}}(t; \tilde{\theta}) - \Vec{X}_{obs}(t)}_2^2 \label{eq:inv-observation-loss-ft}\\
    \mathcal{L}_{\lambda_i} &= \min(0,\vec{\hat{\lambda}}_i - \lambda_{i}^{min})^2-\max(0,\vec{\hat{\lambda}}_i - \lambda_i^{max})^2\label{eq:inv-param-loss-ft}
\end{align}
Compared to the dynamics loss term in \cref{eq:missing-dynamics-loss} from \cref{sec:missing-method}, Equation \cref{eq:inv-dynamics-loss-ft} in this fine-tuning phase encodes the known dynamics for the problem by using the known functional form of the RHS, denoted $\vec{f}_{param}$, with the estimated physical parameter values $\vec{\hat{\lambda}}$ instead of using the observed data directly. The form of the parameter loss terms in \cref{eq:inv-param-loss-ft} penalizes estimates for each parameter outside of the range $\vec{\hat{\lambda}}_i \in [\lambda_{i}^{min}, \lambda_{i}^{max}]$, $i=1, \dots, |\lambda|$. We then sum the individual parameter losses in \cref{eq:inv-loss}. 
Note that the numerical method approximation $\vec{X}_{\textit{NM}}$ is not explicitly used in the loss term for the fine-tuning phase, but due to pre-training the neural network, the numerical approximation implicitly contributes to the learning of $\tilde{\theta}$. \Cref{alg:inverse-ft} provides pseudocode for the neural network fine-tuning phase. In the same way as \cref{sec:missing-method}, we refer to the output of the neural network as the model prediction for this problem. 

\begin{algorithm}[t!]
    \caption{Neural network fine-tuning to estimate unknown physical parameters} \label{alg:inverse-ft}
    \begin{algorithmic}
        \REQUIRE Discretized time steps $t$, initial conditions $\Vec{X}_0$, randomly initialized physical parameters $\Vec{\hat{\lambda}}$, pre-trained neural network parameters $\Vec{\theta}$
        \ENSURE State approximation $\Vec{X}_{\textit{DNN}}$ at time steps $t$, estimated physical parameters $\Vec{\hat{\lambda}}$
        \STATE Define an optimizer \textit{optim}
        \STATE Set $\Tilde{\theta} = [\Vec{\theta}; \Vec{\hat{\lambda}}]$
        \WHILE{\textit{optim} tolerance not met}
        \STATE Reset gradients of \textit{optim}
        \STATE $\Vec{X}_{\textit{DNN}} = \textit{DNN}(t;\tilde{\theta})$
        \STATE Compute required gradients of $\Vec{X}_{\textit{DNN}}$
        \STATE Compute initial condition loss term: $\mathcal{L}_{ic}(\tilde{\theta}) = \frac{1}{N_{ic}}\norm{\Vec{X}_{\textit{DNN}}(t_0;\tilde{\theta})-\Vec{X}_0(t_0)}_2^2$
        \STATE Compute dynamics loss term: $\mathcal{L}_{dyn}(\tilde{\theta}) = \frac{1}{N_{dyn}}\norm{\hat{\vec{f}}_{\textit{DNN}}(t;\tilde{\theta})- \Vec{f}_{param}(t; \vec{\hat{\lambda}})}_2^2$
        \STATE Compute observation loss term: $\mathcal{L}_{obs}(\tilde{\theta}) = \frac{1}{N_{obs}}\norm{\Vec{X}_{\textit{DNN}}(t; \tilde{\theta}) - \Vec{X}_{obs}(t)}_2^2$      
        \STATE Compute parameter loss term $\forall \lambda_i$: $\mathcal{L}_{\lambda_i} = \min(0,\vec{\hat{\lambda}}_i - \lambda_{i}^{min})^2-\max(0,\vec{\hat{\lambda}}_i - \lambda_i^{max})^2$ 
        \STATE $\mathcal{L}(\tilde{\theta}) = \mathcal{L}_{ic}(\tilde{\theta}) + \mathcal{L}_{dyn}(\tilde{\theta}) + \mathcal{L}_{obs}(\tilde{\theta}) + \sum_{i=1}^{|\lambda|} \mathcal{L}_{\lambda_i}$
        \STATE Compute gradients of $\mathcal{L}(\tilde{\theta})$ using backpropagation
        \STATE Update $\tilde{\theta}$ by taking an \textit{optim} step
        \ENDWHILE
    \end{algorithmic}
\end{algorithm}

\subsection{Numerical Experiments for Parameter Estimation}
\label{sec:inverse-results}
In setting up the numerical experiments for parameter estimation, we use the same method as described in \cref{sec:results-methodology}. Similar to the dynamics discovery experiments in \cref{sec:missing-results}, we show results for a suitable choice of numerical method, and in \cref{sec:stiff-inv} we include an additional example comparing performance across different LMM schemes, particularly for a stiff problem requiring implicit solvers. We use a similar approach to generate testing data but compared to \cref{sec:results-methodology}, we only need to randomly select time steps within the domain that the neural networks were not trained on. In addition to the MSE to evaluate the performance of our approach to parameter estimation, we also compare the estimated physical parameter values with the ground truth values used in generating the data by calculating the relative error.

\subsubsection{Example 1: FitzHugh-Nagumo Model}
\label{sec:FNmodel-inv}
We consider the same FitzHugh-Nagumo example as in \cref{sec:FNmodel-missing}. However, for this section, we aim to estimate the unknown physical parameters $\lambda = (a, b, c, z)^T$ in \cref{eq:fn-v,,eq:fn-u}, given both noiseless and 20\% noisy observed data.

\begin{figure}[t!]
    \begin{subfigure}[t]{\textwidth}
        \centering
        \includegraphics[width=0.49\textwidth]{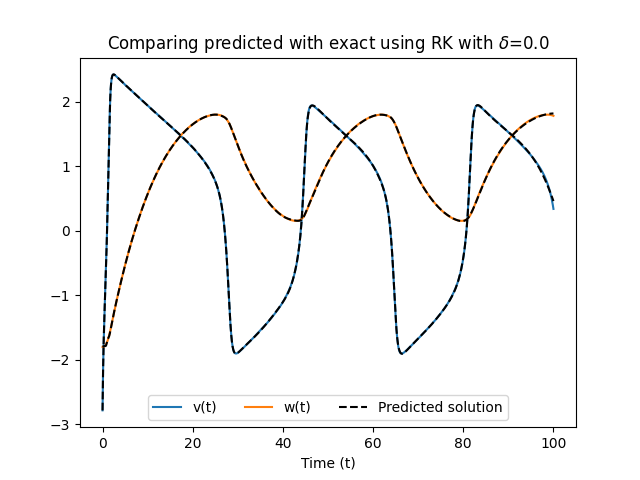}
        \includegraphics[width=0.49\textwidth]{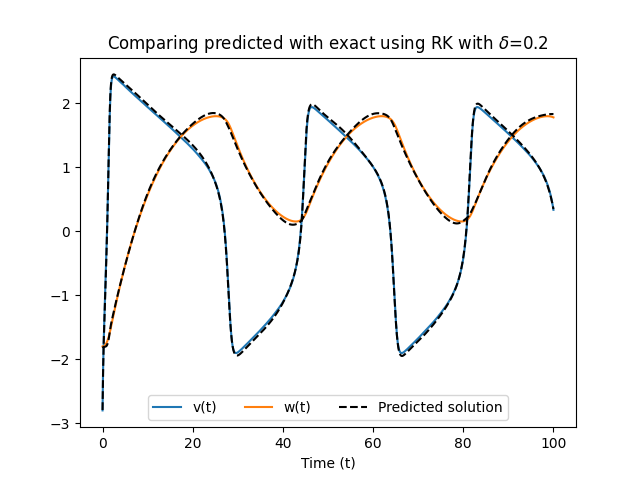}
        \caption{Model predictions of the states $v(t)$ and $w(t)$ using the output of the trained neural network.}
        \label{fig:FN-param-NN}
    \end{subfigure}
    \begin{subfigure}[t]{\textwidth}
        \centering
        \includegraphics[width=0.49\textwidth]{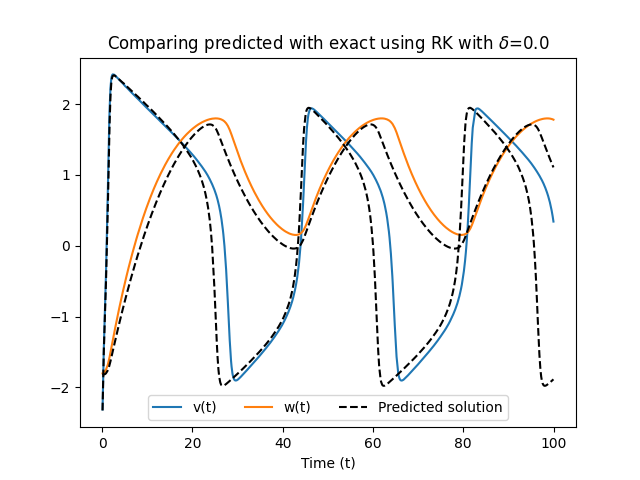}
        \includegraphics[width=0.49\textwidth]{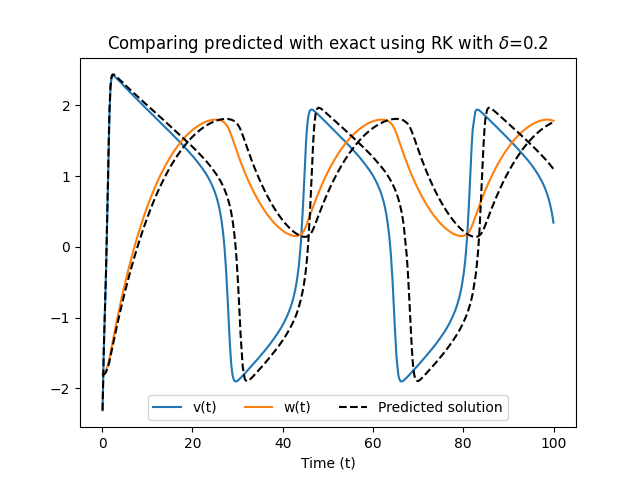}
        \caption{State predictions computed from RK45 using estimated parameter values in \cref{tab:FN-param}.}
        \label{fig:FN-param-estimated}
    \end{subfigure}
    \caption{Parameter estimation predictions of the FitzHugh-Nagumo model states obtained using observed data at various noise levels ($\delta = 0, 0.2$) with RK45.}
    \label{fig:FN-param}
\end{figure}

\paragraph{Results}
We use the same time step and neural network architecture as in \cref{sec:FNmodel-missing} but concatenate the physical parameter estimates to the pre-trained neural network parameters.
\Cref{tab:FN-param} displays the resulting physical parameter estimates and relative errors for each parameter using both RK45 and AB2 methods. \Cref{fig:FN-param} presents the resulting FitzHugh-Nagumo state predictions with RK45 computed in two ways: \cref{fig:FN-param-NN} shows the model prediction of $v(t)$ and $w(t)$ using the output of the trained neural network, while \cref{fig:FN-param-estimated} displays the state prediction obtained by plugging in the estimated physical parameter values from \cref{tab:FN-param} into \cref{eq:fn-v,,eq:fn-u} and solving numerically. \Cref{tab:FN-inv} reports the corresponding MSEs computed using the model predictions and the exact solution described in \cref{sec:FNmodel-inv}.

From these results, we observe model predictions close to exact using the neural network output in \cref{fig:FN-param-NN}, which is reflected in the low MSE values in \cref{tab:FN-inv}. Alternatively, the state predictions in \cref{fig:FN-param-estimated} using the estimated physical parameter values appear slightly out of phase with the ground truth. We note that several of the estimated physical parameter values have somewhat high relative errors reported in \cref{tab:FN-param} compared with the true values used in generating the data. Moreover, there are several estimated parameters that do not change from the initialized value. This phenomenon arises from the fact that we approximate the unknown physical parameters along with the neural network parameters during training. Thus, we observe a compensation for errors in the physical parameter estimates in the estimated neural network parameters $\theta$. Furthermore, this is a point-wise computation of the parameters. We consider ensemble-based methods for uncertainty quantification in this framework in \cref{app:ensemble} and observe closer physical parameter estimates using simple averaging of deep ensemble learning for the parameter estimation problem.

To avoid redundancy, \cref{fig:FN-param} displays results using only RK45 as the numerical scheme, but experiments using AB2 provide similar results. We investigate the importance of the pre-training and fine-tuning steps in \cref{app:inv-training} and conclude that using either training approach individually does not yield the same quality of model predictions compared to the proposed approach combining pre-training on the numerical approximation and fine-tuning for physical parameter learning.

\begin{table}[t!]
    \centering \footnotesize
    \renewcommand{\arraystretch}{1.2}
    \begin{tabular}{|c|c|c|c|c|c|c|}\hline
     & Parameter & True Value & Noise Level & Initial Value & Estimate & Relative Error \\ \hline
    \multirow{8}{*}{RK} & \multirow{2}{*}{$a$} & \multirow{2}{*}{0.7} & 0\% & 0.386 & 0.507 & 0.276 \\ \cline{4-7} 
     & & & 20\% & 0.418 & 0.545 & 0.222 \\ \cline{2-7}
     & \multirow{2}{*}{$b$} & \multirow{2}{*}{0.8} & 0\% & 0.417 & 0.505 & 0.369 \\  \cline{4-7} 
     & & & 20\% & 0.091 & 0.722 & 0.097 \\ \cline{2-7} 
     & \multirow{2}{*}{$c$} & \multirow{2}{*}{12.5} & 0\% & 14.175 & 12.926 & 0.034 \\ \cline{4-7}
     & & & 20\% & 13.084 & 12.955 & 0.036 \\ \cline{2-7}
     & \multirow{2}{*}{$z$} & \multirow{2}{*}{1} & 0\% & 0.572 & 0.814 & 0.186 \\ \cline{4-7}
     & & & 20\% & 1.021 & No change & 0.021 \\ \hline
    \multirow{8}{*}{LMM} & \multirow{2}{*}{$a$} & \multirow{2}{*}{0.7} & 0\% & 0.510 & No change & 0.272 \\ \cline{4-7} 
     & & & 20\% & 0.003 & 0.652 & 0.068 \\ \cline{2-7} 
     & \multirow{2}{*}{$b$} & \multirow{2}{*}{0.8} & 0\% & 0.625 & No change & 0.219 \\ \cline{4-7} 
     & & & 20\% & 0.636 & No change & 0.205 \\ \cline{2-7} 
     & \multirow{2}{*}{$c$} & \multirow{2}{*}{12.5} & 0\% & 13.667 & 12.897 & 0.032 \\ \cline{4-7} 
     & & & 20\% & 13.636 & 12.940 & 0.035 \\ \cline{2-7} 
     & \multirow{2}{*}{$z$} & \multirow{2}{*}{1} & 0\% & 0.876 & No change & 0.124 \\ \cline{4-7}
     & & & 20\% & 0.567 & 0.871 & 0.129 \\ \hline
    \end{tabular}
    \caption{Estimated physical parameter values from the FitzHugh-Nagumo model obtained using noiseless and 20\% noisy observed data with RK (RK45) and LMM (AB2) schemes. Initial values (from pre-training), estimated values, and relative errors are reported to three decimal places. Entries labeled ``No change" have no change up to the reported three decimal places between the pre-training and final estimates.}
    \label{tab:FN-param}
\end{table}

\begin{table}[t!]
    \centering \footnotesize
    \begin{tabular}{|c|ll|ll|}
    \hline
    \multirow{2}{*}{Noise Level} & \multicolumn{2}{c|}{RK} & \multicolumn{2}{c|}{AB2} \\ \cline{2-5} 
     & \multicolumn{1}{c}{$v$} & \multicolumn{1}{c|}{$w$} & \multicolumn{1}{c}{$v$} & \multicolumn{1}{c|}{$w$} \\ \hline
     0\% & 9.50e-05 & 3.21e-05 & 2.45e-04 & 3.12e-05 \\
     20\% & 2.49e-03 & 1.67e-03 & 2.98e-03 & 1.81e-03 \\ \hline
    \end{tabular}
    \caption{Comparing RK and LMM scheme MSEs for FitzHugh-Nagumo parameter estimation model predictions.}
    \label{tab:FN-inv}
\end{table}

\subsubsection{Example 2: Stiff ODE System Modeling Biochemical Kinetics}
\label{sec:stiff-inv}
Suppose we have a system of ODEs modeling the enzymatic reaction between three substances $S_1$, $S_2$, and $S_3$, with concentrations $x_1$, $x_2$, and $x_3$, respectively. The governing equations are given as 
\begin{align}
    \frac{dx_1}{dt} &= \phi(t) - V_1 \frac{x_1}{x_1 + k_1} \label{eq:stiff-ODE-x1}\\[0.2cm]
    \frac{dx_2}{dt} &= V_1 \frac{x_1}{x_1 + k_1} - V_2\frac{x_2}{x_2 + k_2} \label{eq:stiff-ODE-x2}\\[0.2cm]
    \frac{dx_3}{dt} &= V_2\frac{x_2}{x_2 + k_2} - \lambda(x_3 - c_0) \label{eq:stiff-ODE-x3}
\end{align}
where $\phi(t) = A_0 + A(t-t_0) \exp\{-(t-t_0)/\tau\}$ is an input of substance $S_1$ and $\lambda$, $c_0$, $A_0$, $A$, $t_0$, and $\tau$ are known parameters; a similar model was considered as a stiff test problem in \cite{Arnold2013, Arnold2014}. 
For parameter estimation, given $j \in \{1, 2\}$, we aim to determine the maximum reaction rates $V_j$ and affinity constants $k_j$ of the Michaelis-Menten model terms \cite{michaelismenten1913}. We note that this system is stiff for certain parameter values.

To apply the proposed method to this problem, we take $\lambda = 1.5$, $c_0 = 0.5$, $\tau = 0.5$, $A = e/\tau$, and $A_0 = 0.1$. The true values for the unknown physical parameters are as follows: $V_1 = 1.0$, $V_2 = 0.5$, $k_1 = 0.1$, and $k_2 = 0.5$. We choose $[0,0,0]$ as the initial condition and $\Delta t = 0.01$. For this example, the parameter loss terms from \eqref{eq:inv-param-loss-ft} use $\vec{\hat{\lambda}}_i \in [0.5 \lambda_i, 1.5 \lambda_i]$, where each $\lambda_i$ corresponds to the true parameter values defined previously. 

To evaluate the performance of the proposed methods across different LMM schemes, we conduct several numerical experiments comparing the resulting model predictions for the parameter estimation problem in Section \ref{sec:inverse-method}. In particular, we compare results across the 2-step schemes AB2 (explicit), AM2 (implicit), and BDF2 (implicit) on this stiff ODE example. For the example at hand, we note that 2-step methods provide a reasonable balance between computational efficiency and accurate results.

\begin{figure}[t!]
    \centering
    \includegraphics[width=0.9\textwidth]{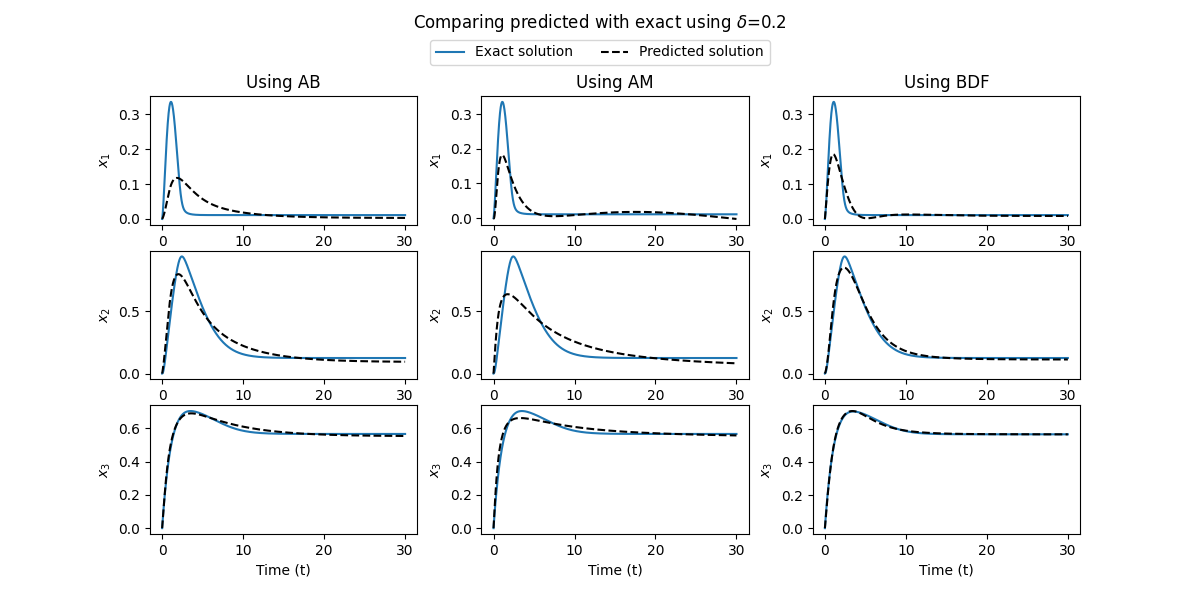}
    \caption{Parameter estimation model predictions of the stiff ODE system using various LMM 2-step schemes with 20\% noisy observed data.}
    \label{fig:ODE-paramLMM-comp-inv}
\end{figure}

\paragraph{Results}
We use an architecture of 4 hidden layers with 64 nodes each and train according to the method described in \cref{sec:inverse-method}.
When comparing the model predictions for the parameter estimation approach in \cref{fig:ODE-paramLMM-comp-inv}, we see variation in accuracy across the three numerical schemes considered. This is most evident in the predictions of the initial spikes of the $x_1$ and $x_2$ concentrations for $t \in [0, 10]$---while none of the models well capture the magnitude of the $x_1$ spike, BDF2 shows the best predictive performance overall across all three components, perhaps unsurprisingly due to the stiffness of the system. 
For the learned physical parameters in \cref{tab:ODE-LMMcomp-inv-param}, we observe similar results as discussed in \cref{sec:FNmodel-inv}, with some unchanged values and fairly high relative errors. We particularly note a high relative error for the $k_1$ parameter using AB2 because of a parameter estimate far from the true value; however, the initial parameter value decreased significantly through the fine-tuning phase. We posit with more epochs of training, the relative error could decrease, as discussed for future work in \cref{sec:discussion}.

\begin{table}[t!]
    \centering\footnotesize
    \begin{tabular}{|l|ccc|cc|}
    \hline
    \multirow{2}{*}{} & \multicolumn{3}{c|}{MSE} & \multicolumn{2}{c|}{Training Time (min)} \\ \cline{2-6} 
     & \multicolumn{1}{c|}{$x_1$} & \multicolumn{1}{c|}{$x_2$} & $x_3$ & \multicolumn{1}{c|}{Pre-training} & Fine-tuning \\ \hline
    AB2 & \multicolumn{1}{c|}{2.44e-03} & \multicolumn{1}{l|}{3.18e-03} & 1.61e-04 & \multicolumn{1}{c|}{1.318} & 3.654 \\ \hline
    AM2 & \multicolumn{1}{c|}{9.43e-04} & \multicolumn{1}{l|}{9.80e-03} & 4.90e-04 & \multicolumn{1}{c|}{2.385} & 5.215 \\ \hline
    BDF2 & \multicolumn{1}{l|}{7.87e-04} & \multicolumn{1}{l|}{6.23e-04} & 1.90e-05 &  \multicolumn{1}{c|}{1.226} & 4.076 \\ \hline
    \end{tabular}
    \caption{Comparing various LMM 2-step scheme MSEs and computational time for the stiff ODE system parameter estimation problem.}
    \label{tab:ODE-LMMcomp-inv}
\end{table}

\begin{table}[t!]
    \centering\footnotesize
    \begin{tabular}{|c|c|c|c|c|c|}\hline
     & Parameter & True Value & Initial Value & Estimated Value & Relative Error \\ \hline
      \multirow{4}{*}{AB2} & $V_1$ & 1.0 & 0.148 & 0.382 & 0.618 \\ 
     & $V_2$ & 0.5 & 0.776 & 0.750 & 0.500  \\  
     & $k_1$ & 0.1 & 0.944 & 0.476 & 3.762 \\ 
     & $k_2$ & 0.5 & 0.454 & No change & 0.092 \\ \hline
      \multirow{4}{*}{AM2} & $V_1$ & 1.0 & 0.054 & 0.486 & 0.514 \\
     & $V_2$ & 0.5 & 0.932 & 0.750 & 0.500  \\ 
     & $k_1$ & 0.1 & 0.692 & 0.174 & 0.740  \\ 
     & $k_2$ & 0.5 & 0.719 & No change & 0.438 \\ \hline
     \multirow{4}{*}{BDF2} & $V_1$ & 1.0 & 0.455 & 0.561 & 0.439 \\
     & $V_2$ & 0.5 & 0.311 & No change & 0.378  \\ 
     & $k_1$ & 0.1 & 0.182 & 0.107 & 0.068  \\ 
     & $k_2$ & 0.5 & 0.839 & 0.630 & 0.262 \\ \hline
    \end{tabular}
    \caption{Comparing estimated physical parameter values from the stiff ODE system with 20\% noisy observed data for various LMM 2-step schemes. Entries labeled ``No change" have no change up to the reported three decimal places between the pre-training and final estimates.}
    \label{tab:ODE-LMMcomp-inv-param}
\end{table}

\section{Discussion and Conclusions}
\label{sec:discussion}
This work presents two novel approaches combining deep learning and numerical methods to address dynamics discovery and parameter estimation in deterministic dynamical systems. In these missing physics problems, we augment neural network training with numerical ODE solvers to approximate unknown system states and physical parameter values from noisy observed data. Implementing the proposed methods on a suite of test problems provides empirical evidence that encoding a loss function with numerical methods, such as RK45 and different LMMs, leads to promising predictions of unknown system dynamics and physical parameters, even with significant Gaussian noise corrupting the observed data. Our models demonstrate strong generalization when applied to oscillatory and chaotic problems. 

Results using RK and LMM families of methods are comparable, given appropriate choices of time step, order, and neural network architecture. As illustrated in \cref{sec:stiff-inv}, the choice of implicit LMMs such as the BDF family results in reasonable accuracy for stiff problems, but these methods require solving a nonlinear optimization problem at every step in addition to the non-convex optimization problem when training the neural network. From \cref{tab:FN-missing,,tab:FN-inv,,tab:lorenz,,tab:heat}, we observe similar MSE values between RK and LMM when comparing the model prediction with the exact solutions.

We did not observe significant differences in the resulting model predictions and parameter estimates due to the addition of noise in the observed data, suggesting that our proposed approaches are robust to noise. However, the test examples used simulated noisy data generated from known solutions since we considered well-known problems in computational science. In future work, we will apply these methods to real-world datasets that contain significant noise (as well as modeling errors) and are more representative of missing physics in dynamical systems. Moreover, we will investigate methods to account for this noise, not only within the testing data but also the neural network itself. Extending the work in \cref{app:ensemble}, we will explore additional techniques for modeling uncertainty and noise within the proposed architecture. 

When estimating the system physical parameters, we observe that the proposed method implementing a pre-training and fine-tuning phase recovers values reasonably close to the exact, even when trained on data perturbed with significant Gaussian noise. Moreover, the method provides accurate predictions of the system states using the estimated parameters $\hat{\lambda}$, even when the parameters include some approximation errors, as demonstrated with the low MSEs in \cref{tab:FN-inv,,tab:lorenz,,tab:heat}. In some cases, the final physical parameter estimates include relatively large approximation errors or remain unchanged during the neural network training for both the noiseless and noisy cases. This arises from the fact that the unknown physical parameters are optimized along with the neural network parameters. When training the neural network, the prediction improves because of the tuned neural network parameters $\theta$ even though the learned physical parameters $\hat{\lambda}$ may differ from their exact values. Physical parameters that do not update may be unindentifiable from the given data or randomly initialized to reasonable approximations already in the pre-training phase. Future work will address different neural network architectures and parameter updating strategies to better approximate the physical parameter values.

In \cref{sec:FNmodel-missing,,app:weight}, we investigated how loss term weighting affected the resulting model predictions for dynamics discovery. While there was a large discrepancy in model predictions for excluding the dynamics loss term, the results for various nonzero weights did not have a significant effect. In future work, we will explore additional hyperparameters associated with loss term weighting. Further, in combining neural networks and numerical ODE solvers, we note that the choice of hyperparameters involving the spatial and temporal discretization schemes, numerical method orders, and neural network architectures play a significant role in our proposed method, with appropriate choices of these settings leading to promising results. We will further investigate what constitutes a good choice of these hyperparameters and how to choose them optimally in future work.

\vspace{2.5mm}
\noindent\textbf{Data Availability Statement:} Code and data to reproduce the results in this paper will be made available upon reasonable request.

\vspace{2.5mm}
\noindent\textbf{Declaration of Competing Interest:} None.


\bibliography{paper_refs}{}

\newpage
\appendix
\appendixsectionformat
\appendixfiguretablenumbering 
\section{Loss Term Weighting}
\label{app:weight}
\begin{figure}[b!]
    \centerline{\includegraphics[width=0.5\textwidth]{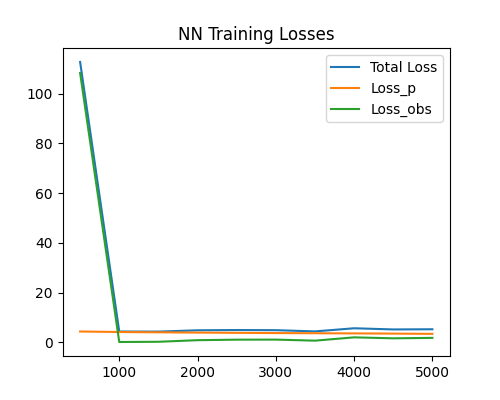} \includegraphics[width=0.5\textwidth]{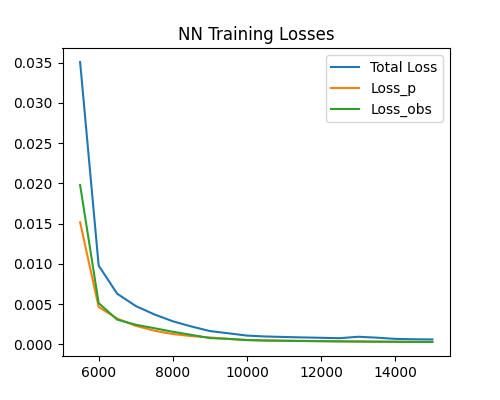}}
    \caption{FitzHugh-Nagumo training loss function for dynamics discovery (left) and parameter estimation fine-tuning (right) using RK and $\delta = 0.0$.}
    \label{fig:FN-loss-plots}
\end{figure}

The examples in \cref{sec:missing-results,,sec:inverse-results} demonstrate the effect of physics-informed loss functions on model learning when the observation and dynamics loss terms from \cref{sec:missing-method,,sec:inverse-method} are weighted equally. However, depending on the known physics of a particular problem, we may weight the observation or dynamics loss term more or less heavily to account for missing information. For example, in the dynamics discovery problem with an unknown RHS, the loss function relied heavily on observed, noisy trajectories. Comparably, in the parameter estimation problem with a known RHS but unknown physical parameters, the loss function relied more heavily on the dynamics loss term. \cref{fig:FN-loss-plots} illustrates this difference. Thus, we investigate how loss term weighting can affect the results on the FitzHugh-Nagumo example outlined in Section \ref{sec:FNmodel-missing} with noiseless data. More concretely, for weight $W \in \R$, we train on the following loss function:
\begin{equation}
    \mathcal{L}(\theta) = \mathcal{L}_{ic}(\theta) + W \cdot \mathcal{L}_{dyn}(\theta) + \mathcal{L}_{obs}(\theta).
\end{equation}

\begin{figure}[t!]
    \centerline{ \includegraphics[width=0.33\textwidth]{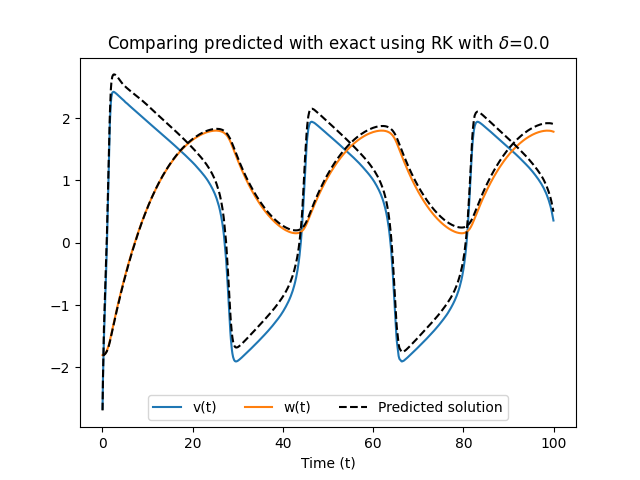}\includegraphics[width=0.33\textwidth]{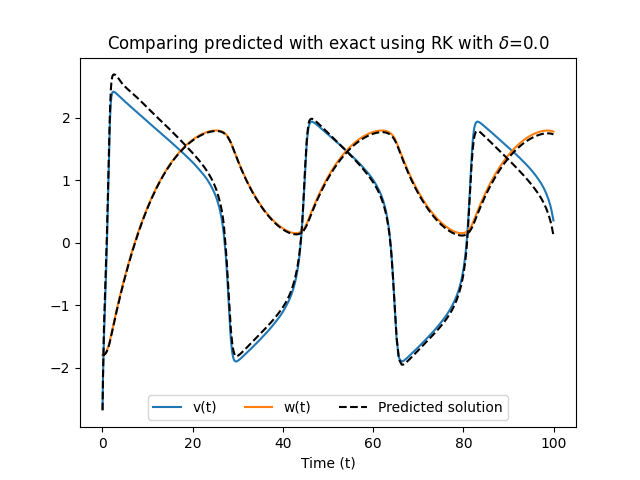} \includegraphics[width=0.33\textwidth]{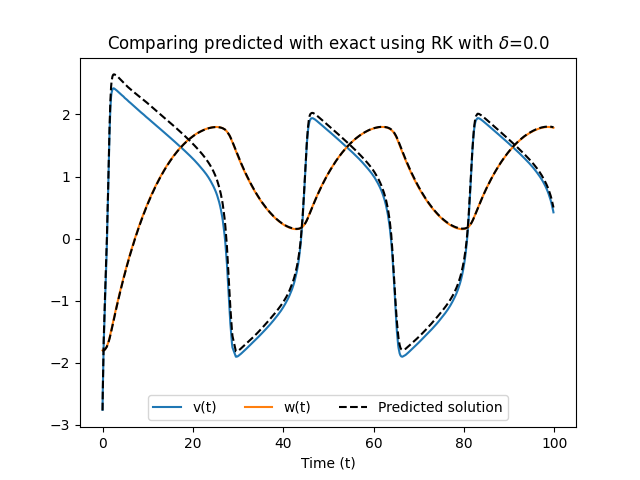}}
    \caption{Dynamics discovery model predictions of the FitzHugh-Nagumo model obtained using $W \in \{10, 100, 1000\}$. }
    \label{fig:missing-weight10}
\end{figure}

\begin{figure}[t!]
    \centerline{\includegraphics[width=0.33\textwidth]{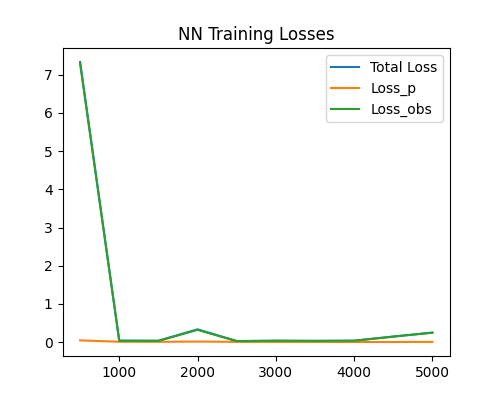} \includegraphics[width=0.33\textwidth]{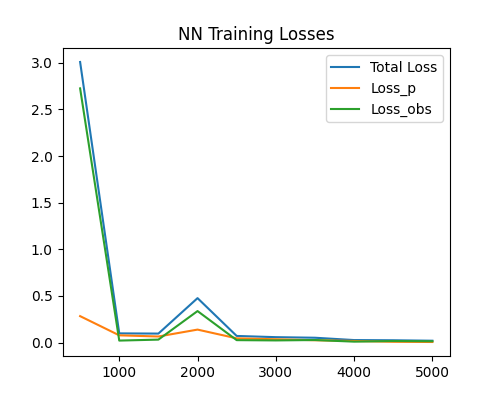} \includegraphics[width=0.33\textwidth]{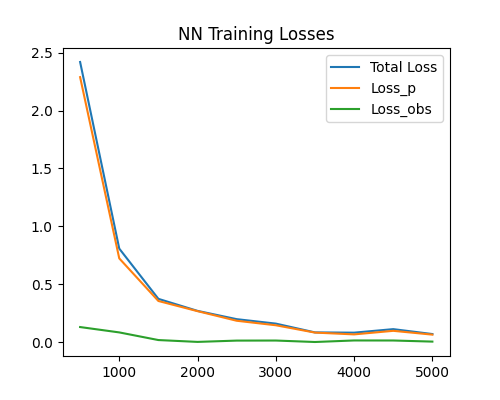}}
    \caption{FitzHugh-Nagumo training loss function for $W \in \{10, 100, 1000\}$. }
    \label{fig:missing-weight10-loss}
\end{figure}

\begin{figure}[t!]
    \centerline{\includegraphics[width=0.33\textwidth]{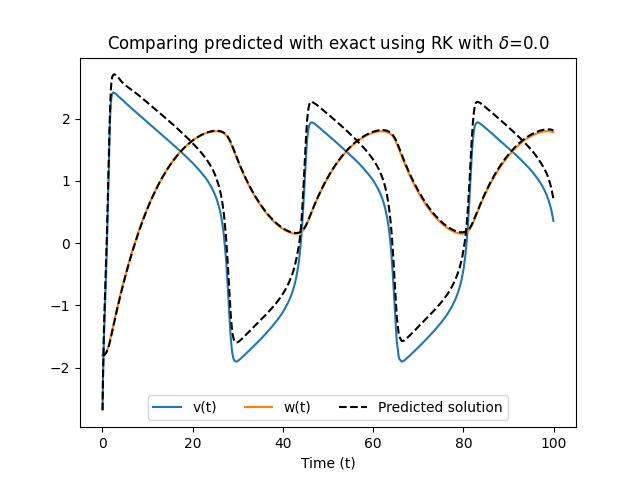} \includegraphics[width=0.33\textwidth]{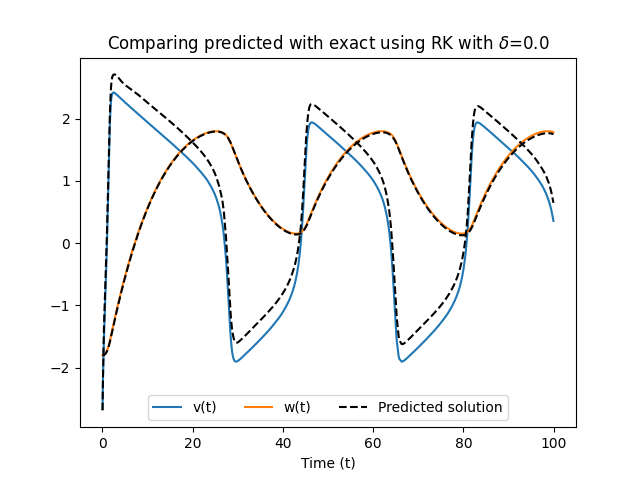} \includegraphics[width=0.33\textwidth]{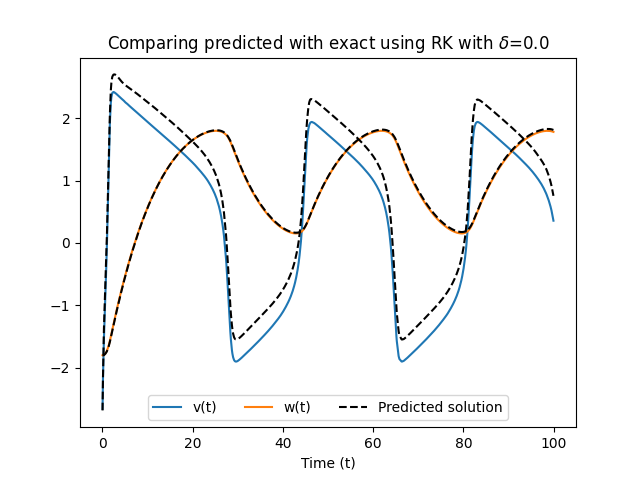}}
    \caption{Dynamics discovery model predictions of the FitzHugh-Nagumo model obtained using $W \in \{0.1, 0.01, 0.001\}$. }
    \label{fig:missing-weight0.1}
\end{figure}

\begin{figure}[t!]
    \centerline{\includegraphics[width=0.33\textwidth]{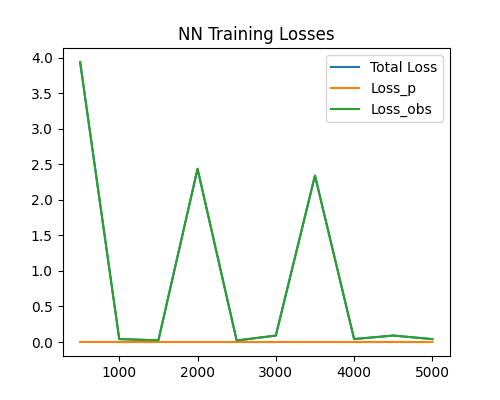} \includegraphics[width=0.33\textwidth]{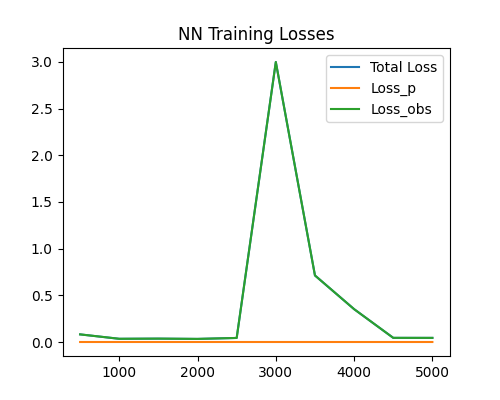} \includegraphics[width=0.33\textwidth]{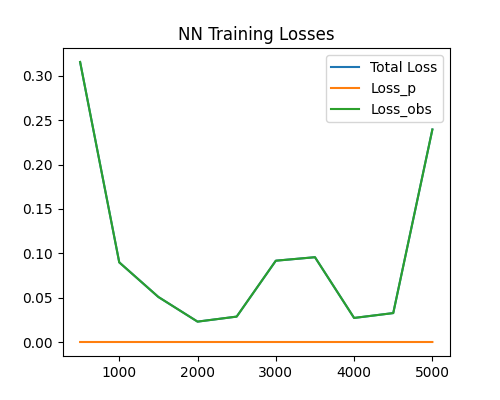}}
    \caption{FitzHugh-Nagumo training loss function for $W \in \{0.1, 0.01, 0.001\}$. }
    \label{fig:missing-weight0.1-loss}
\end{figure}

To investigate how loss term weighting may affect the results for the dynamics discovery problem, we implement the exact same setup for noiseless training as in \cref{sec:FNmodel-missing} and choose various values for $W$ in the loss function. When we choose $W > 1$, the resulting plots in \cref{fig:missing-weight10} show comparable model predictions as well as similar MSE values in \cref{tab:missing-weight}. In the context of the dynamics discovery problem, this can be interpreted as biasing toward the physics of the problem, which is inherently unknown. As shown in \cref{fig:missing-weight10-loss}, the training losses indicate reasonable optimization, even as $W$ increases. Alternatively, if we choose $W \in [0, 1)$, we observe a significant detrimental effect on model predictions, as seen in both \cref{fig:missing-weight0.1} and \cref{tab:missing-weight}. Compared to the results obtained in \cref{sec:FNmodel-missing} with $W=1$, the model predictions appear farther from exact and the MSEs are larger. Furthermore, the training loss plots in \cref{fig:missing-weight0.1-loss} do not show a gradual minimization of the loss terms.
As discussed in \cref{sec:FNmodel-missing}, when we set $W=0$, the model predictions indicate that the neural network cannot learn the dynamics of the system from observations alone. When compared with the loss function proposed in \cref{eq:missing-loss}, we have empirical evidence that the encoding of the dynamics loss term contributes to the improved model predictions for the FitzHugh-Nagumo dynamics discovery model.
With the results discussed in this appendix, we have empirical evidence that relying solely on observed trajectories may result in poor model predictions without also accounting for some known physics in the form of numerical approximations. 

\begin{table}[t!]
    \centering\footnotesize
    \begin{tabular}{|c|cc|}
    \hline
    \multirow{2}{*}{$W$} & \multicolumn{2}{c|}{MSE} \\ \cline{2-3} 
     & $v$ & $w$ \\ \hline
    0     & 7.57e+00 & 2.79e+01 \\
    0.001 & 1.25e-01 & 2.75e-04 \\ 
    0.01  & 8.87e-02 & 4.14e-04 \\
    0.1   & 1.02e-01 & 2.68e-04 \\
    1     & 1.34e-02 & 8.28e-05 \\
    10    & 4.25e-02 & 4.13e-03 \\
    100   & 1.84e-02 & 7.54e-04 \\
    1000  & 3.94e-02 & 3.14e-05 \\ \hline
    \end{tabular}
    \caption{Comparing MSEs for FitzHugh-Nagumo model predictions with varying weights on $\mathcal{L}_{dyn}$.}
    \label{tab:missing-weight}
\end{table}

\section{Pre-training and Fine-tuning Set Up}
\label{app:inv-training}
As detailed in \cref{sec:inverse}, the method for physical parameter estimation utilizes a pre-training step to initialize the neural network parameters $\theta$ before fine-tuning the neural network for estimating physical parameter values. Without this additional step, the task of approximating system states and physical parameters becomes much more difficult and the resulting predictions are not as accurate. 

\begin{figure}[t!]
    \centering
    \begin{subfigure}[b]{0.5\textwidth}
        \centering
        \includegraphics[width=\textwidth]{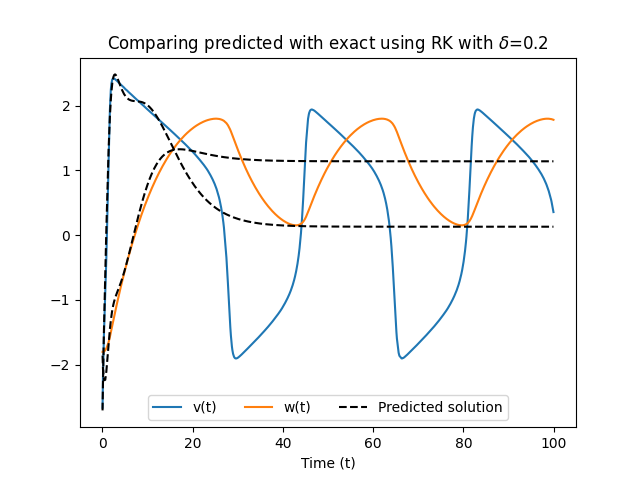}
        \caption{MSE for states $v(t)$ and $w(t)$ are 1.51e+00 \\and 3.17e-01, respectively.}
        \label{fig:fn-paramRK-noise-noPRE}
    \end{subfigure}%
    \begin{subfigure}[b]{0.5\textwidth}
        \centering
        \includegraphics[width=\textwidth]{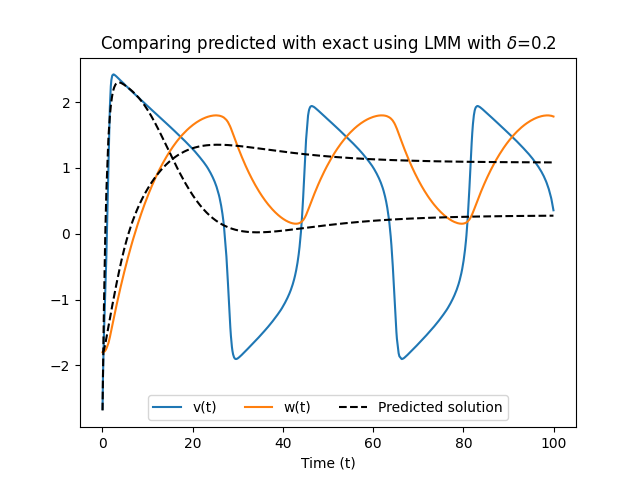}
        \caption{MSE for states $v(t)$ and $w(t)$ are 1.48+00 \\and 2.95e-01, respectively.}
        \label{fig:fn-paramLMM-noise-noPRE}
    \end{subfigure}
    \caption{Parameter estimation model predictions of FitzHugh-Nagumo model with 20\% noisy data for RK (RK45) and LMM (AB2) schemes without pre-training.}
    \label{fig:FN-param-noise-noPRE}
\end{figure}

To highlight the importance of this pre-training phase, \cref{fig:FN-param-noise-noPRE} and \cref{tab:FN-param-noisy-noPRE} display the results of applying the same neural network architecture described in \cref{sec:FNmodel-inv} but removing the pre-training step. That is, we only use the fine-tuning training described in \cref{sec:inv-ft} with the augmented parameter vector $\tilde{\theta} = (\theta; \vec{\hat{\lambda}})^T$, where the neural network parameters $\theta$ are randomly initialized instead of coming from pre-training. As seen in \cref{fig:FN-param-noise-noPRE}, the predicted solutions for $v(t)$ and $w(t)$ are nowhere near as close to the exact solutions with 20\% noisy data, losing the dynamics in both components around time $t\in(10,20)$, which is also reflected in the relatively high MSEs compared to the results that included the pre-training step (see \cref{tab:FN-inv}). 
When comparing the physical parameter estimates, several of the estimated values obtained without pre-training in \cref{tab:FN-param-noisy-noPRE} do not update from their initial values, similar to the case with pre-training. We note that while the relative errors for some parameter estimates are smaller compared to the corresponding results with pre-training (see \cref{tab:FN-param}), the pre-trained neural network parameters allow the network to better learn the system dynamics via the numerical solution of the system, resulting in better model predictions than without pre-training. 

\begin{figure}[t!]
    \centering
    \begin{subfigure}[b]{0.5\textwidth}
        \centering
        \includegraphics[width=\textwidth]{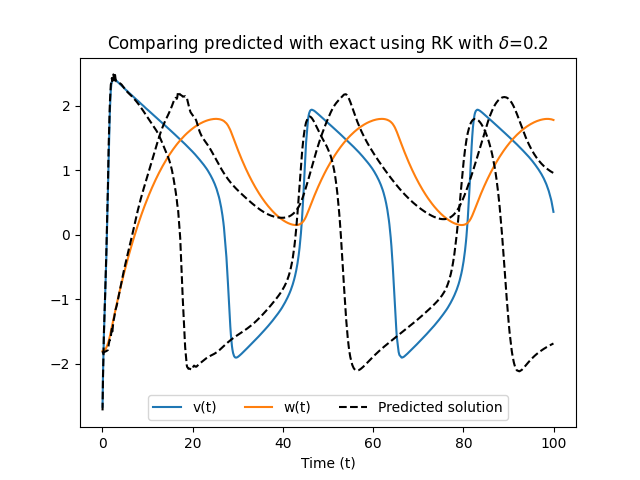}
        \caption{MSE for states $v(t)$ and $w(t)$ are 2.84e+00 \\and 2.72e-01, respectively.}
        \label{fig:fn-paramRK-noise-noFT}
    \end{subfigure}%
    \begin{subfigure}[b]{0.5\textwidth}
        \centering
        \includegraphics[width=\textwidth]{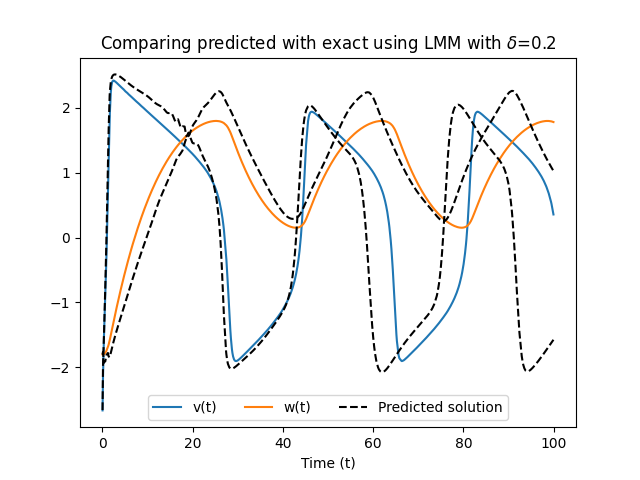}
        \caption{MSE for states $v(t)$ and $w(t)$ are 1.41e+00 \\and 1.79e-01, respectively.}
        \label{fig:fn-paramLMM-noise-noFT}
    \end{subfigure}
    \caption{Parameter estimation model predictions of FitzHugh-Nagumo model with 20\% noisy data for RK (RK45) and LMM (AB2) schemes without fine-tuning.}
    \label{fig:FN-param-noise-noFT}
\end{figure}

The fine-tuning phase is also a necessity for this proposed approach. \cref{fig:FN-param-noise-noFT} display the results of applying the same neural network architecture described above but removing the fine-tuning step. That is, we only use the pre-training described in \cref{sec:inv-pt} with randomly initialized physical parameter values that are not learned in the process of neural network training. As seen in \cref{fig:FN-param-noise-noFT}, the predicted solutions for both state components are far from exact although the generally oscillatory behavior is somewhat recovered due to the numerical approximation $\Vec{X}_{\textit{NM}}$ utilized in the loss function. The MSEs for these predictions are also comparable to those when we trained only with the fine-tuning loss functions, which are both significantly higher than the MSEs computed in \cref{tab:FN-inv} for parameter estimation at the same noise level. Thus, we illustrate the necessity of both pre-training and fine-tuning to achieve reasonable predictions for the state components and physical parameter estimates.

\begin{table}[t!]
    \centering\footnotesize
    \renewcommand{\arraystretch}{1.2}
    \begin{tabular}{|c|c|c|c|c|c|}\hline
     & Parameter & True Value & Initial Value & Estimated Value & Relative Error \\ \hline
    \multirow{4}{*}{RK} & $a$ & 0.7 & 0.252 & 0.585 & 0.165 \\ \cline{2-6} 
     & $b$ & 0.8 & 0.083 & 0.642 & 0.197 \\ \cline{2-6} 
     & $c$ & 12.5 & 10.048 & 12.405 & 0.008 \\ \cline{2-6} 
     & $z$ & 1 & 1.147 & No change & 0.147 \\ \hline
    \multirow{4}{*}{LMM} & $a$ & 0.7 & 0.981 & 0.853 & 0.218 \\ \cline{2-6} 
     & $b$ & 0.8 & 0.566 & No change & 0.293 \\ \cline{2-6} 
     & $c$ & 12.5 & 11.181 & 12.435 & 0.005 \\ \cline{2-6} 
     & $z$ & 1 & 1.018 & No change & 0.018 \\ \hline
    \end{tabular}
    \caption{Estimated physical parameter values from the FitzHugh-Nagumo model obtained using 20\% noisy observed data with RK (RK45) and LMM (AB2) schemes without pre-training. Randomly initialized values, estimated values, and relative errors are reported to three decimal places. Entries labeled ``No change" have no change up to the reported three decimal places between the initial value and final estimates.}
    \label{tab:FN-param-noisy-noPRE}
\end{table}

\section{Ensemble Averaging}
\label{app:ensemble}
Deep ensemble methods aggregate the predictions of multiple point estimate neural networks to quantify uncertainty that may arise from approximating the system of interest. To obtain a more accurate final prediction with error estimates, we consider a simple averaging approach to ensemble learning which gives the combined output as 
\begin{equation}
    x_{avg}(t) = \frac{1}{M}\sum_{i=1}^{M}x_i(t),
\end{equation}
where each $x_i(t)$ represents the predicted output from the $i$-th neural network and $M$ is the total number of neural networks in the ensemble.

We investigate how deep ensemble learning may provide better uncertainty estimates on the FitzHugh-Nagumo example outlined in \cref{sec:FNmodel-missing,,sec:FNmodel-inv}. We train multiple models using the approaches described in \cref{sec:missing,,sec:inverse} to obtain a mean predicted state across the ensemble of trained neural networks. We train each neural network on different training data within the domain and tested on the same test set of random time steps within the domain for consistency across models. 

\begin{figure}[t!]
    \centerline{\includegraphics[width=0.33\textwidth]{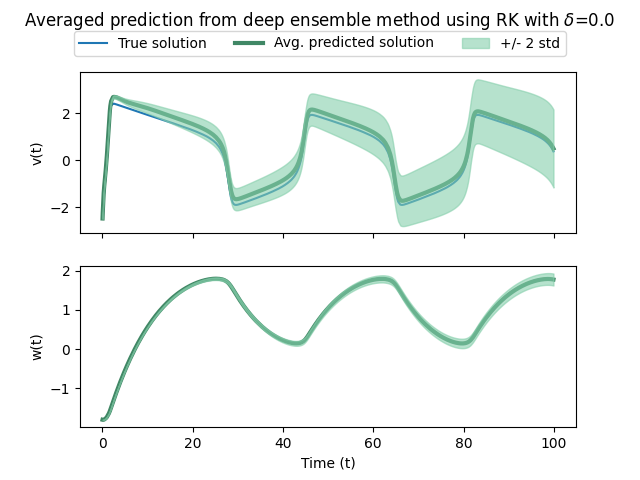}\includegraphics[width=0.33\textwidth]{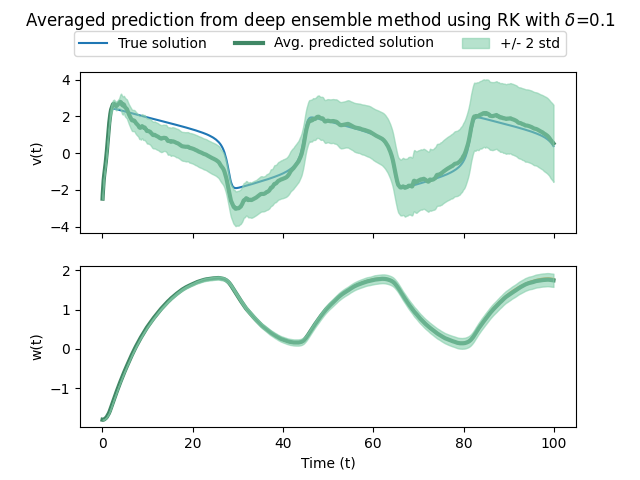}\includegraphics[width=0.33\textwidth]{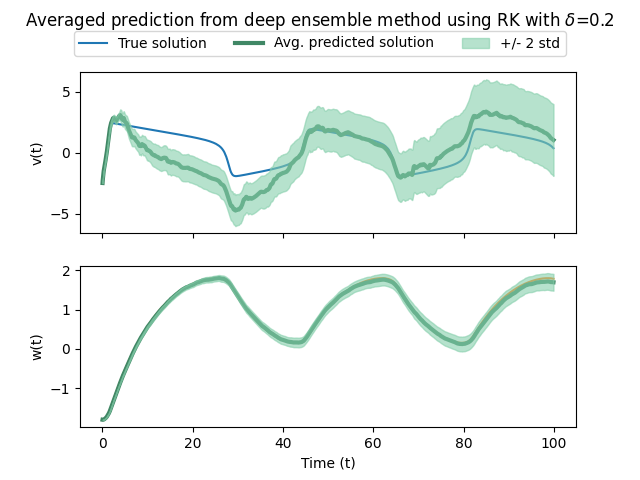}}
    \caption{Deep ensemble averaging predictions after training 50 models for dynamics discovery.}
    \label{fig:missing-ensemble}
\end{figure}

\paragraph{Dynamics Discovery}
We implement the same details described in \cref{sec:FNmodel-missing} and train 50 models independently on noiseless data, averaging the resulting state predictions. From \cref{fig:missing-ensemble}, we obtain uncertainty estimates based on the variability in model training. The MSEs in \cref{tab:FN-ensemble} are slightly higher compared to those computed for a point-estimate model in \cref{tab:FN-missing,,tab:FN-inv} at the same noise-level because deep ensemble learning introduces added variability. Furthermore, we observe that the predicted uncertainty appears to be underestimated, particularly from $t\in [8, 30]$ in the $v(t)$ component when $\delta = 0.2$ in \cref{fig:missing-ensemble}. This could be due to how deep ensemble methods capture noise in the model not the data itself. In \cref{sec:lorenz-missing}, we observed state predictions drifting from the exact solutions and recovering some of the dynamics at a later time. With deep ensemble learning, this drifting behavior also appears to have an effect on the result in \cref{fig:missing-ensemble}.

\begin{figure}[t!]
    \centerline{\includegraphics[width=0.33\textwidth]{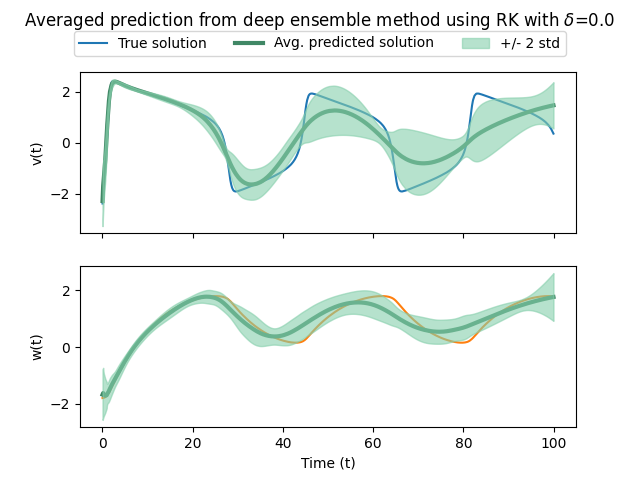}\includegraphics[width=0.33\textwidth]{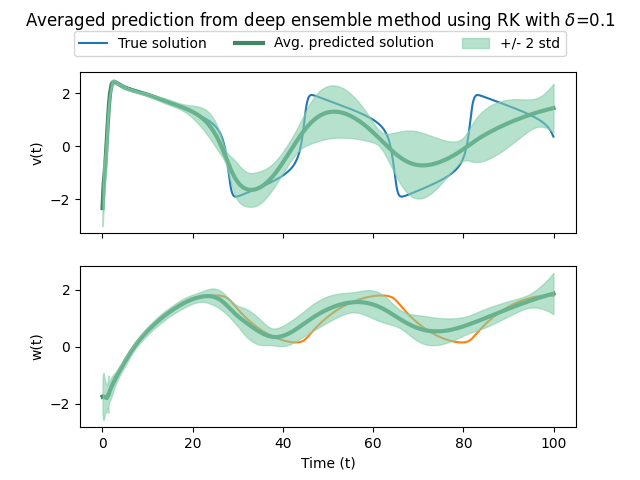}\includegraphics[width=0.33\textwidth]{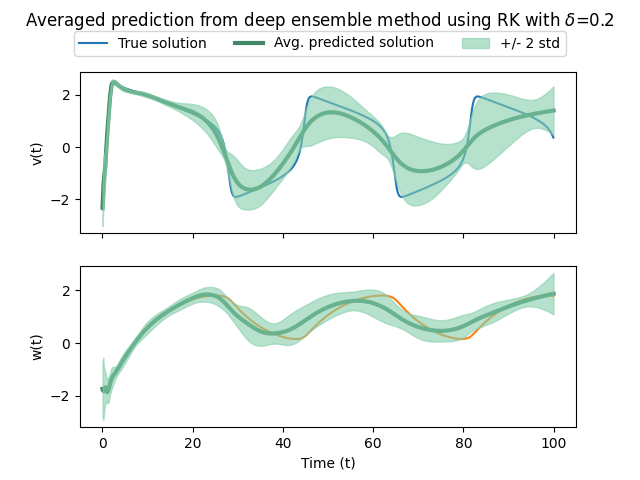}}
    \caption{Deep ensemble averaging predictions after training 50 models for parameter estimation.}
    \label{fig:param-ensemble}
\end{figure}

\paragraph{Parameter Estimation}
Similar to the dynamics discovery problem, we implement the same details from \cref{sec:FNmodel-inv} on noiseless data. We train 50 models and aggregate both the resulting state predictions and physical parameter estimates across the ensemble models. We observe a degregation in mean state prediction of $v(t)$ after $t=25$ in \cref{fig:param-ensemble} due to the fact that we average the result of 50 models, but the uncertainty bounds account for this error. Additionally, \cref{tab:FN-param-ensemble} displays the resulting average physical parameter estimates and relative errors for each parameter. In comparison to the parameter estimates from the point-estimate model in \cref{tab:FN-param}, these aggregate physical parameter estimates appear closer to exact as the relative errors are smaller. Similar to the dynamics discovery ensemble problem, the MSEs in \cref{tab:FN-ensemble} for parameter estimation are much higher compared to those in \cref{tab:FN-inv} for the point-estimate model. 

\begin{table}
    \centering \footnotesize
    \renewcommand{\arraystretch}{1.2}
    \begin{tabular}{|c|c|c|c|c|c|c|}\hline
     & Parameter & True Value & Noise Level & Initial Value & Estimate & Relative Error \\ \hline
    \multirow{12}{*}{RK} & \multirow{3}{*}{$a$} & \multirow{3}{*}{0.7} & 0\% & 0.553 & 0.675 & 0.036 \\ \cline{4-7} 
     & & & 10\% & 0.685 & 0.709 & 0.013\\ \cline{4-7}
     & & & 20\% & 0.819 & 0.650 & 0.072\\ \cline{2-7} 
     & \multirow{3}{*}{$b$} & \multirow{3}{*}{0.8} & 0\% & 0.480 & 0.691 & 0.136 \\ \cline{4-7} 
     & & & 10\% & 0.915 & 0.721 & 0.099\\ \cline{4-7}
     & & & 20\% & 0.756 & 0.723 & 0.096\\ \cline{2-7} 
     & \multirow{3}{*}{$c$} & \multirow{3}{*}{12.5} & 0\% & 13.351 & 12.524 & 0.002 \\ \cline{4-7} 
     & & & 10\% & 13.426 & 12.579 & 0.006\\ \cline{4-7}
     & & & 20\% & 10.541 & 12.527 & 0.002\\ \cline{2-7} 
     & \multirow{3}{*}{$z$} & \multirow{3}{*}{1} & 0\% & 0.815 & 1.053 & 0.053 \\ \cline{4-7} 
     & & & 10\% & 0.908 & 0.973 & 0.027\\ \cline{4-7}
     & & & 20\% & 1.123 & 1.020 & 0.020\\\hline 
     \multirow{12}{*}{LMM} & \multirow{3}{*}{$a$} & \multirow{3}{*}{0.7} & 0\% & 0.033 & 0.729 & 0.042 \\ \cline{4-7} 
     & & & 10\% & 0.742 & 0.669 & 0.044\\ \cline{4-7}
     & & & 20\% & 0.631 & 0.668 & 0.045 \\ \cline{2-7} 
     & \multirow{3}{*}{$b$} & \multirow{3}{*}{0.8} & 0\% & 0.503 & 0.660 & 0.175 \\ \cline{4-7} 
     & & & 10\% & 0.565 & 0.724 & 0.095\\ \cline{4-7}
     & & & 20\% & 0.390 & 0.623 & 0.221\\ \cline{2-7} 
     & \multirow{3}{*}{$c$} & \multirow{3}{*}{12.5} & 0\% & 14.717 & 12.531 & 0.003 \\ \cline{4-7} 
     & & & 10\% & 11.235 & 12.525 & 0.002 \\ \cline{4-7}
     & & & 20\% & 14.058 & 12.481 & 0.001\\ \cline{2-7} 
     & \multirow{3}{*}{$z$} & \multirow{3}{*}{1} & 0\% & 1.079 & 1.039 & 0.039 \\ \cline{4-7} 
     & & & 10\% & 1.346 & 0.962 & 0.038 \\ \cline{4-7}
     & & & 20\% & 1.295 & 0.970 & 0.030\\ \hline
    \end{tabular}
    \caption[Estimated physical parameter values from the FitzHugh-Nagumo model obtained from observed data with $\delta = 0, 0.1, 0.2$ using RK and LMM schemes]{Estimated physical parameter values from the FitzHugh-Nagumo model obtained from observed data with $\delta = 0, 0.1, 0.2$ using RK (RK45) and LMM (AB2) schemes. Initial values (from one instance of pre-training), aggregate estimated values, and aggregate relative errors are reported to three decimal places.}
    \label{tab:FN-param-ensemble}
\end{table}

\begin{table}
    \centering \footnotesize
    \begin{tabular}{|l|c|ll|ll|}
    \hline
    \multirow{2}{*}{} & \multirow{2}{*}{Noise Level} & \multicolumn{2}{c|}{RK} & \multicolumn{2}{c|}{LMM} \\ \cline{3-6} 
     &  & \multicolumn{1}{c}{$v$} & \multicolumn{1}{c|}{$w$} & \multicolumn{1}{c}{$v$} & \multicolumn{1}{c|}{$w$} \\ \hline
    \multirow{3}{*}{Dynamics Discovery} & 0\% & 4.52e-02 & 1.04e-04 & 5.51e-02 & 1.92e-04\\
     & 10\% & 2.95e-01 & 3.20e-04 & 2.80e-01 & 5.36e-04 \\
     & 20\% & 2.18e+00 & 1.54e-03 & 2.09e+00 & 1.58e-03 \\ \hline
    \multirow{3}{*}{Parameter Estimation} & 0\% & 3.77e-01 & 6.63e-02 & 3.19e-01 & 5.86e-02 \\
     & 10\% & 4.06e-01 & 6.89e-02 & 3.65e-01 & 7.62e-02 \\ 
     & 20\% & 3.02e-01 & 7.00e-02 & 3.65e-01 & 7.61e-02 \\ \hline
    \end{tabular}
    \caption[Comparing MSEs for FitzHugh-Nagumo ensemble model predictions]{Comparing RK and LMM scheme MSEs for FitzHugh-Nagumo ensemble model predictions. For this example, we use BDF2 for dynamics discovery and AB2 for parameter estimation as our LMMs.}
    \label{tab:FN-ensemble}
\end{table}

\end{document}